\documentclass{article}

\usepackage[preprint]{neurips_2026}
\usepackage{float}
\usepackage{microtype}
\PassOptionsToPackage{final,pagebackref=true}{hyperref}
\usepackage{hyperref}
\usepackage{url}
\usepackage{booktabs}
\usepackage{graphicx}
\usepackage{capt-of}
\usepackage{subcaption}
\usepackage{algorithm}
\usepackage{listings}
\usepackage[table]{xcolor}  \usepackage{multirow} 
\usepackage{xspace}

\hypersetup{
	colorlinks=true,       
	linkcolor=blue,        
	citecolor=blue,        
	filecolor=magenta,     
	urlcolor=blue
}

\lstdefinestyle{pyalg}{
  language=Python,
  basicstyle=\ttfamily\scriptsize,
  keywordstyle=\color{blue!65!black},
  commentstyle=\color{teal!60!black},
  showstringspaces=false,
  columns=fullflexible,
  keepspaces=true,
  breaklines=true,
  frame=none,
  aboveskip=0.2em,
  belowskip=0em
}

\usepackage{amsmath,amsfonts,amsthm}
\usepackage[capitalize,noabbrev]{cleveref}
\theoremstyle{remark}

\newcommand{\nameshort}{\textsc{Repr-Align}\xspace}

\title{Don't Retrain---Align: Adapting Autoregressive LMs to Diffusion LMs via Representation Alignment}

\author{
Fred Zhangzhi Peng\thanks{Correspondence to: \texttt{zhangzhi.peng@duke.edu}}\\
Duke University\\
\And
Alexis Fox\\
Duke University\\
\AND
Anru R. Zhang\\
Duke University\\
\And
Alexander Tong\\
AITHYRA
}
\newcommand{\mask}{\ensuremath{\langle \mathrm{M} \rangle}}

\begin{document}

\maketitle

\begin{abstract}
Diffusion language models (DLMs) have recently demonstrated capabilities that complement standard autoregressive (AR) models, particularly in non-sequential generation and bidirectional editing. Although recent work has shown that pretrained autoregressive checkpoints can be converted into diffusion language models, existing recipes primarily transfer parameters through continued denoising training with objective- and attention-level modifications. We instead ask whether the internal representation geometry learned by next-token prediction can be explicitly preserved during AR$\rightarrow$DLM conversion. We hypothesize that much of the semantic structure learned by AR pretraining can transfer across generation orders, and thus DLM training should be viewed as relearning the decoding path rather than relearning language representations. To investigate this, we introduce \nameshort, a \emph{Representation Alignment} objective that adapts a bidirectional masked diffusion model to reuse representations from a pretrained AR model of identical architecture. Concretely, we align the hidden states of the DLM to the frozen AR model at every layer using cosine similarity, while optimizing the standard masked denoising objective. This simple alignment---with no adapters and no architectural changes beyond the attention mask---yields up to \emph{4}$\times$ training acceleration in our setting and is particularly effective in low-data regimes.
Our results suggest that linguistic representations are universal regardless of generation order, and representation alignment could be a new go-to technique for training diffusion language models. Code is available at \url{https://github.com/pengzhangzhi/Open-dLLM}.

\end{abstract}

\begin{figure}[H]
    \centering
    \begin{subfigure}[]{0.45\linewidth}
      \centering
      \includegraphics[width=\linewidth,height=0.175\textheight,keepaspectratio]{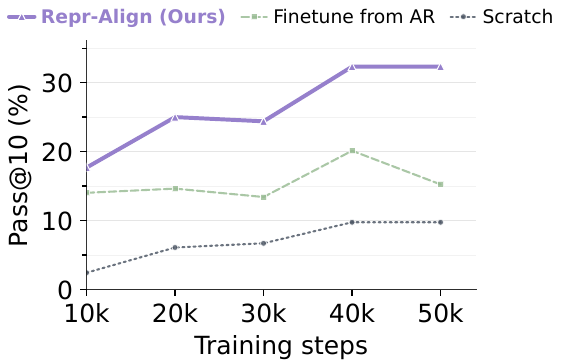}
      \caption{Adaptation speed.}
      \label{fig:humaneval-pass10}
    \end{subfigure}
    \hfill
    \begin{subfigure}[]{0.45\linewidth}
      \centering
      \includegraphics[width=\linewidth,height=0.175\textheight,keepaspectratio]{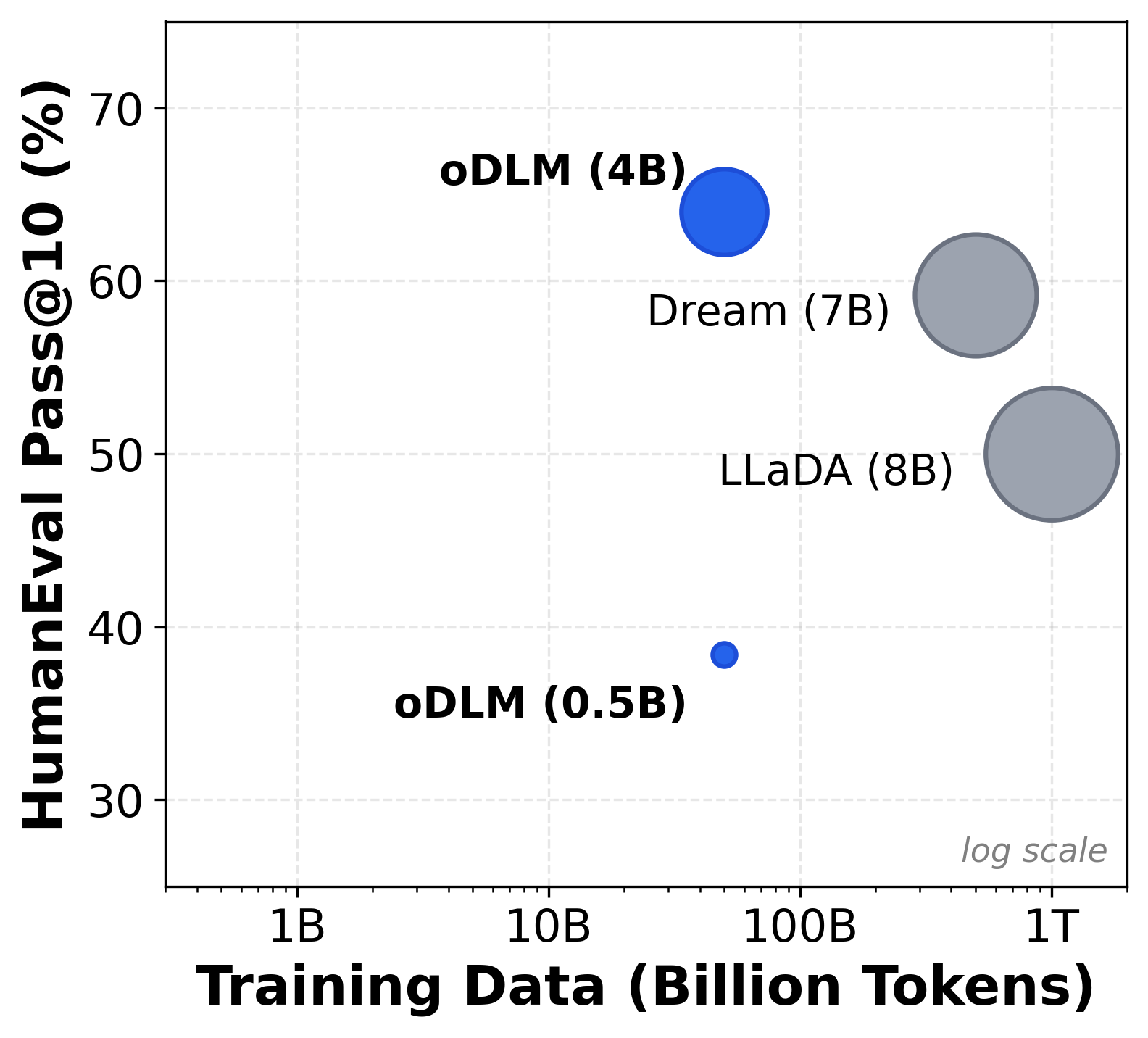}
      \caption{Public DLM frontier.}
      \label{fig:modelscompare}
    \end{subfigure}
    \caption{
  \textbf{Don't retrain---align.}
  \textbf{Left:} \nameshort consistently accelerates AR$\rightarrow$DLM adaptation on HumanEval pass@10, outperforming both AR fine-tuning and scratch training throughout early conversion.
  \textbf{Right:} The resulting oDLM achieves a favorable HumanEval pass@10 versus training-data trade-off among public DLMs.
  }
    \label{fig:teaser}
  \end{figure}

\section{Introduction}
\label{sec:introduction}
\begin{figure}[t]
  \centering
  \includegraphics[width=0.78\linewidth]{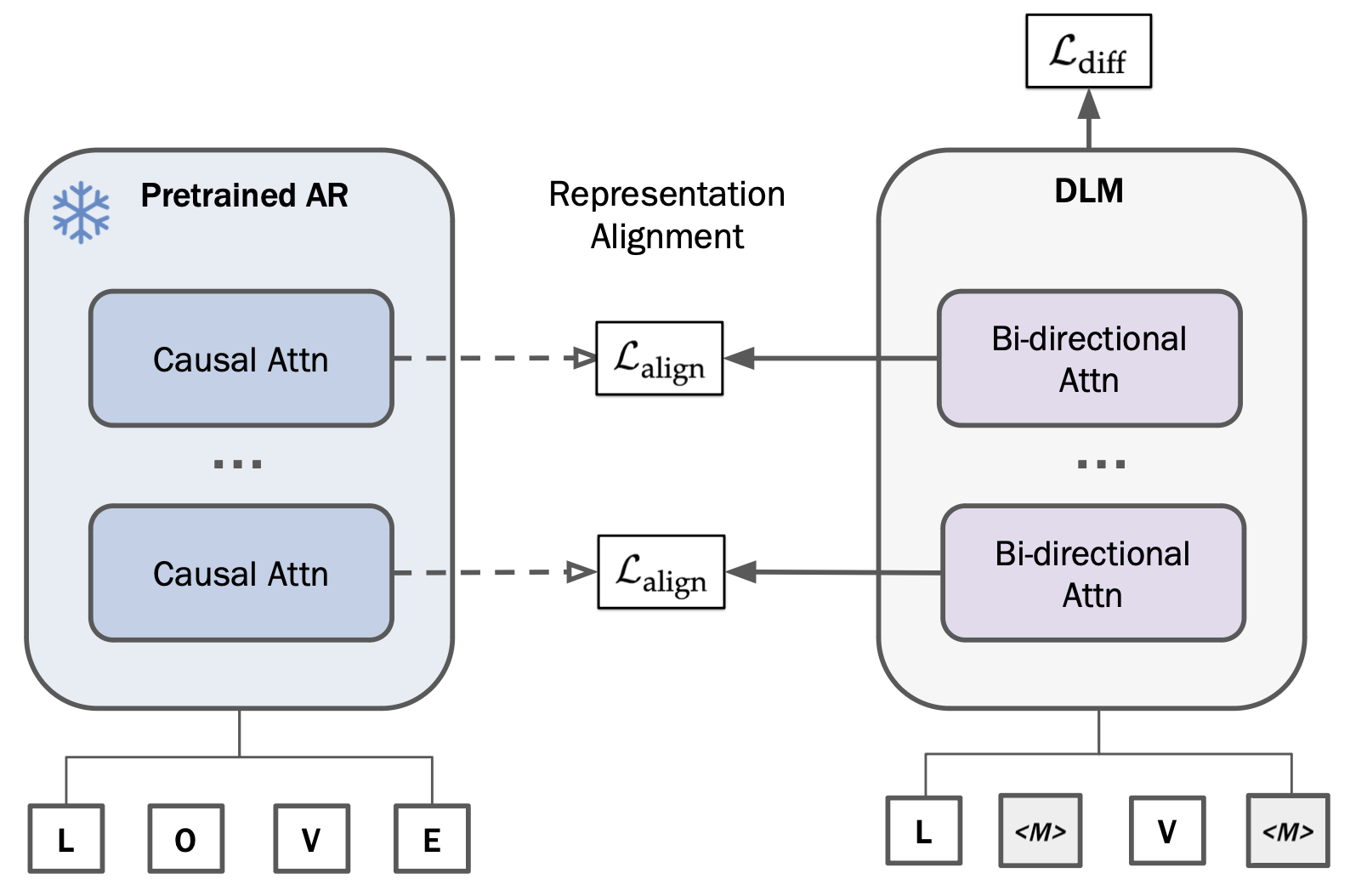}
  \caption{Overview of our method \nameshort: we adapt a pretrained autoregressive (AR) transformer into a masked diffusion language model (DLM) by switching to bidirectional attention and training with a masked denoising objective, while anchoring layer-wise hidden states to a frozen AR backbone.}
  \label{fig:method}
\end{figure}

The dominant paradigm in large-scale language modeling has long been autoregressive (AR) sequence modeling. By factoring the joint probability distribution as a product of conditional probabilities, models such as GPT and Qwen have demonstrated strong general-purpose generation capabilities~\citep{radford2019language,achiam2023gpt,Touvron2023Llama2O}. Recently, diffusion language models (DLMs) have emerged as an alternative formulation for text generation, spanning continuous diffusion over embeddings, discrete absorbing-state diffusion, likelihood-based diffusion LMs, masked diffusion LMs, and large-scale DLMs~\citep{li2022diffusionlm,Austin2021StructuredDD,Gulrajani2023LikelihoodBasedDL,mdlm,nie2025llada,dream2025}. By framing generation as any-order decoding~\citep{pmlr-v37-sohl-dickstein15,ho2020denoising,yang2019xlnet,ghazvininejad2019maskpredict}, DLMs naturally support non-left-to-right behaviors such as infilling and iterative refinement~\citep{mdlm,Gulrajani2023LikelihoodBasedDL,Chang_2022_CVPR}.
Despite these advantages, scaling DLMs remains expensive. In theory DLMs learn $L!$ paths to generate a sequence vs. one left-to-right generation in ARs, and thus require $L$-times more compute. While several recent methods reduce this cost by initializing from pretrained AR checkpoints or converting AR models into DLMs~\citep{gong2025scalingdiffusionlanguagemodels,dream2025,efficientdlm2025}, existing conversion recipes largely reuse AR parameters through continued denoising training, attention-pattern modifications, or sampling conventions. They do not explicitly constrain the converted DLM to preserve the internal representation geometry of the AR model.

In this paper, we question the need to treat ARs and DLMs as two disjoint paradigms. We start from a simple view: the hard part of language generation is learning language representations---the semantic and syntactic structure of the data---not committing to a particular generation order. Autoregressive pretraining has already learned strong internal features that organize this structure. If so, training a diffusion language model should not require relearning language representations from scratch. Instead, the remaining work is mainly mechanical: adapt these existing features to an iterative any-order decoder. This reframes DLM training from representation learning to an alignment problem, where we reuse the AR backbone for the representations and train the diffusion mechanism to operate in the same feature space but with any-order generation.

To test this hypothesis, we apply \emph{Representation Alignment}~\citep{yu2025repa,singh2025irepa,wu2025representation,jiang2025sra} for the first time for a minimalist adaptation of a pretrained AR transformer into a masked diffusion language model (\Cref{fig:method}). Our setup uses two models with identical architecture: (i) a pretrained AR model with causal attention, and (ii) the same architecture initialized from the AR weights but with bidirectional attention. During training, we randomly mask a sequence and optimize the DLM to predict the masked tokens. In parallel, we feed the clean sequence into the frozen AR model (teacher-forced under causal attention) and extract its hidden states at each layer. Because the two networks share the same layer structure and hidden sizes, we can directly align their intermediate representations via a layer-wise cosine-similarity loss, without introducing adapters or additional parameters. Intuitively, the AR model provides a stable representational anchor, and diffusion training is reduced to learning an any-order decoding mechanism that operates in that anchored feature space.
The method is designed to change as little as possible, so that any gains can be attributed to the reuse of AR features rather than to architectural modifications or heavy fine-tuning.

Our experiments support the reuse-through-alignment view: representation alignment turns AR$\rightarrow$DLM conversion into a largely mechanical adaptation problem rather than relearning linguistic representations.
On HumanEval, alignment improves conversion quality and the gains grow with model size (\Cref{fig:repralign_curve}), increasing pass@10 from 24.9 to 31.0 at 0.6B and from 31.1 to 40.5 at 1.7B.
Beyond quality, alignment enables substantially cheaper conversion through selective training (\Cref{fig:freezing}) and remains effective under a 0.8B-token \emph{tiny} subset (\Cref{fig:tiny}).
As a scale-up validation, we train a 4B oDLM using the same representation-preserving conversion recipe.
Against Dream-7B, a strong public DLM that also builds on AR initialization, oDLM achieves a better HumanEval-family pass@10 trade-off (\Cref{fig:modelscompare,tab:code-generation}): it improves HumanEval and HumanEval+ pass@10 by 2.39 and 2.40 points, respectively, despite using fewer parameters and a substantially lighter data-and-compute budget.
\paragraph{Contributions.}
Our contributions are summarized as follows:
\begin{itemize}
    \item We identify representation preservation as a missing ingredient in AR$\rightarrow$DLM conversion. 
    Instead of merely initializing from an AR checkpoint, we explicitly anchor the DLM student to the frozen AR model's layer-wise hidden-state geometry during masked denoising training.

    \item We introduce a simple representation-preserving conversion recipe that requires no adapters or architectural changes beyond switching from causal to bidirectional attention. 
    Across model scales, representation alignment improves conversion quality and sample efficiency, with larger gains at larger model sizes.

    \item We show that AR$\rightarrow$DLM conversion is not inherently data- or parameter-update hungry. 
    With representation alignment, training on a 0.8B-token subset can outperform training on the full 50B-token stream under the same step budget, and freezing embeddings and MLP blocks improves throughput by up to $\sim$2$\times$ without degrading quality.

    \item As a scale-up validation, we train a 4B oDLM using the same recipe. 
    Compared with Dream-7B, a strong public DLM that also leverages AR initialization, oDLM improves HumanEval and HumanEval+ pass@10 by 2.39 and 2.40 points, respectively, while using a smaller backbone and a substantially lighter data-and-compute budget.
\end{itemize}

\section{Related Work}
\label{sec:related_work}
\paragraph{Diffusion language models.}
Diffusion language models span continuous diffusion over embeddings, discrete categorical diffusion, likelihood-based diffusion LMs, and masked diffusion LMs~\citep{li2022diffusionlm,Austin2021StructuredDD,Gulrajani2023LikelihoodBasedDL,mdlm,nie2024scalingmaskeddiffusionmodels}. Recent large-scale systems such as LLaDA and Dream show that masked diffusion can support instruction following, reasoning, and code generation at billion-parameter scale~\citep{nie2025llada,dream2025}. These advances make DLMs a serious alternative to autoregressive generation, but competitive DLMs still require substantial diffusion-specific optimization. We provide a more detailed discussion of DLM formulations in \Cref{app:rw_dlm}.

\paragraph{Adapting autoregressive models to diffusion language models.}
A growing line of work avoids training DLMs from scratch by converting pretrained AR checkpoints into denoising models~\citep{gong2025scalingdiffusionlanguagemodels,dream2025,dreamcoder2025,efficientdlm2025,xue2025anyordergpt}. These methods establish that AR checkpoints are strong initializations, but they primarily adapt the objective, masking process, attention pattern, or sampling convention. Our work instead explicitly preserves the AR model's internal representation geometry by aligning a bidirectional DLM student to a frozen same-architecture AR teacher, as formalized in \Cref{sec:method_alignment}. \Cref{app:rw_ar_to_dlm,app:rw_any_order} give a fuller comparison with AR$\rightarrow$DLM conversion, any-order generation, iterative masked decoding, and path-planning methods.

\paragraph{Representation Alignment for Generative Models.}
Representation alignment has recently accelerated diffusion training by matching generative-model hidden states to representations from strong pretrained encoders~\citep{yu2025repa,singh2025irepa,wu2025representation,jiang2025sra}. Our setting differs in both teacher and purpose: the teacher is not an external encoder, but the exact AR model being converted, with the same tokenizer, architecture, hidden size, and initialization as the DLM student. Alignment therefore acts as representation preservation during mechanism adaptation, rather than feature import from another model; \Cref{eq:align_loss} gives the exact objective. \Cref{app:rw_repr_alignment} expands this distinction.

\section{Method}
\label{sec:method}

We study the question: \emph{are representations learned by next-token prediction sufficient for masked denoising generation?}

To isolate this factor, we keep the architecture fixed and change only what is necessary for diffusion-style denoising. Our method instantiates two transformers with identical parameterization and dimensionality: a pretrained autoregressive (AR) model with causal attention, and a masked diffusion model with bidirectional attention. We then train the diffusion model with the standard masked prediction objective, while adding a single layer-wise representation alignment loss to reuse the AR model's internal features. Alg.~\ref{alg:ar2dlm_align} summarizes the overall conversion procedure. No adapters or auxiliary modules are introduced.
The four design choices below map directly to the experimental setup in \Cref{sec:setup} and the ablations in \Cref{sec:ablations}.

\begin{algorithm}[H]
\caption{\nameshort AR$\rightarrow$DLM conversion with layer-wise representation alignment.}
\label{alg:ar2dlm_align}
\begin{lstlisting}[style=pyalg]
# f_AR: frozen AR teacher (causal attention), params theta_AR
# f_D : DLM student (bidirectional attention), params theta
# theta <- theta_AR; f_D differs only by attention mask (causal -> bidir)

theta = theta_AR
freeze(f_AR); f_AR.eval()

for x in data_stream:                           # x: [B, n] clean tokens
    r = Uniform(0, 1)                           # mask ratio (per sequence)
    M = sample_positions(x, r)
    x_tilde = x.clone(); x_tilde[M] = mask_id   # corrupted input

    # student denoising pass (bidirectional)
    logits, H_D = f_D(x_tilde, bidir=True, output_hidden_states=True)

    # teacher anchor pass (causal) on clean x
    with no_grad():
        _, H_AR = f_AR(x, causal=True, output_hidden_states=True)

    # masked denoising loss (shift logits by one token;)
    loss_diff = CE( shift_logits_1(logits)[M], x[M] )

    # layer-wise cosine alignment on hidden states
    loss_align = (1 - cos(H_D, H_AR)).mean()

    loss = loss_diff + lambda_align * loss_align
    step_optimizer(theta, loss)
\end{lstlisting}
\end{algorithm}

\subsection{Two models, same architecture}
\label{sec:method_models}
Let $x=(x_1,\dots,x_n)\in\mathcal{V}^n$ be a token sequence, and let $\mathcal{V}$ include a special mask token \mask.
We define:
(i) an autoregressive transformer $f_{\mathrm{AR}}(\cdot;\theta_{\mathrm{AR}})$ with a \emph{causal} attention mask, pretrained by next-token prediction; and
(ii) a diffusion transformer $f_{\mathrm{D}}(\cdot;\theta)$ with \emph{bidirectional} attention.
Crucially, $f_{\mathrm{AR}}$ and $f_{\mathrm{D}}$ share the same layer structure and hidden size $d$; the only architectural difference is the attention mask. We keep $\theta_{\mathrm{AR}}$ frozen throughout training. In practice, we initialize $\theta$ from $\theta_{\mathrm{AR}}$ and switch the attention mask from causal to bidirectional, so that any gains can be attributed to mechanism adaptation rather than new capacity.

Let $h^{(\ell)}_{\mathrm{AR}}(x)\in\mathbb{R}^{n\times d}$ and $h^{(\ell)}_{\mathrm{D}}(x)\in\mathbb{R}^{n\times d}$ denote the hidden states at layer $\ell\in\{1,\dots,L\}$ for the two models.
This matched-conversion design is evaluated under the fixed-budget protocol in \Cref{sec:setup,sec:results_alignment}.

\subsection{Masked diffusion objective}
\label{sec:method_mdm}
We train $f_{\mathrm{D}}$ as a masked denoiser.
We sample a mask set $M\subseteq \{1,\dots,n\}$ and construct a corrupted input $\tilde{x}$ by replacing $x_i$ with \mask{} for $i\in M$.
The diffusion model predicts the masked tokens conditioned on $\tilde{x}$, and we optimize cross-entropy only on masked positions:
\begin{equation}
\mathcal{L}_{\mathrm{diff}}(\theta)
=
\mathbb{E}_{x,M}\left[
\sum_{i\in M}
\mathrm{CE}\!\left(p_{\theta}(\cdot\mid \tilde{x})_i,\; x_i\right)
\right],
\label{eq:diff_loss}
\end{equation}
where $p_{\theta}(\cdot\mid \tilde{x})_i$ is the diffusion model's predicted distribution at position $i$.
This denoising objective is the shared training loss used across the baseline and aligned models in \Cref{sec:setup}.

\paragraph{Shift convention.}
\label{sec:method_shift}
To match the next-token indexing convention inherited from the AR checkpoint, we apply the standard shift operation used in AR$\rightarrow$DLM adaptation: the student produces logits at positions $i$, which are shifted by one position before computing the denoising cross-entropy and during sampling.
We use this same shift convention for both the baseline and aligned models; implementation details are provided in \Cref{app:model_conversion}.

\subsection{Layer-wise representation alignment}
\label{sec:method_alignment}
To reuse AR features, we run a frozen autoregressive teacher on the \emph{clean} sequence $x$ under causal attention, and anchor the diffusion student---which consumes the corrupted $\tilde{x}$ under bidirectional attention---to the teacher's hidden states.
Because the two networks share identical architecture and hidden dimensionality, we align their layer-wise representations directly without adapters.

Let $h^{(\ell)}_{\mathrm{AR}}(x)\in\mathbb{R}^{n\times d}$ denote the teacher hidden states at layer $\ell$ given clean $x$, and $h^{(\ell)}_{\mathrm{D}}(\tilde{x})\in\mathbb{R}^{n\times d}$ denote the student hidden states given corrupted $\tilde{x}$.
We minimize a cosine distance loss:
\begin{equation}
\mathcal{L}_{\mathrm{align}}(\theta)
=
\frac{1}{L}\sum_{\ell=1}^{L}
\mathbb{E}_{x,M}\left[
\frac{1}{|\mathcal{I}|}\sum_{i\in\mathcal{I}}
\left(
1
-
\cos\!\Big(
h^{(\ell)}_{\mathrm{D}}(\tilde{x})_i,\;
\mathrm{stopgrad}\big(h^{(\ell)}_{\mathrm{AR}}(x)_i\big)
\Big)
\right)
\right],
\label{eq:align_loss}
\end{equation}
where $\mathcal{I}$ is the aligned position set (default: all positions), and the teacher is run in evaluation mode with stop-gradient.
\emph{Intuitively, the teacher provides a stable representational coordinate system induced by next-token pretraining, and the student learns a denoising mechanism that operates within this coordinate system.}
The choice of cosine distance, aligned layers, and alignment strength is ablated in \Cref{sec:ablations}, while the exact hidden-state tuple and masking positions used for alignment are specified in \Cref{app:repr_alignment_details}.

\subsection{Training objective}
\label{sec:method_objective}
The full objective is a weighted sum:
\begin{equation}
\mathcal{L}(\theta)
=
\mathcal{L}_{\mathrm{diff}}(\theta)
+
\lambda\,\mathcal{L}_{\mathrm{align}}(\theta),
\label{eq:full_loss}
\end{equation}
with a scalar $\lambda$ (set to 10 by default) controlling the strength of the anchor.
We optimize $\theta$ while keeping $\theta_{\mathrm{AR}}$ fixed.
Overall, this procedure changes as little as possible: the architecture is shared, the pretrained AR representations are preserved, and diffusion training is reduced to learning an any-order denoising mechanism that operates within an already-formed feature space. In addition to the standard masked diffusion model loss, we include the PAPL~\citep{peng2026planner} loss from prior work as part of the default DLM training recipe, with weight 1.
\Cref{sec:results_alignment,sec:results_scaling,sec:ablations} test whether the alignment term in \Cref{eq:full_loss} improves conversion quality and how sensitive it is to $\lambda$.

\section{Experiments}
\label{sec:experiment}

\subsection{Experimental Setup}
\label{sec:setup}
\paragraph{AR-to-DLM conversion setting.}
We study efficient adaptation from an autoregressive (AR) causal language model to a masked diffusion language model (DLM) using Qwen3 checkpoints~\citep{qwen3technicalreport} at three scales (0.6B, 1.7B, and 4B parameters). For each scale, we treat the original causal model as a frozen teacher and initialize the DLM student from the same checkpoint, changing only the attention mask from causal to bidirectional.
This setup instantiates the same-architecture construction in \Cref{sec:method_models}.
\paragraph{Training data.}
We train on the Nemotron-SFT-Code dataset~\citep{nvidia2025nvidianemotronnano2} (approximately 50B tokens and 70M sequences). This release is a large synthetically generated and curated SFT-style corpus covering STEM, academic, reasoning, and multilingual instruction data, and includes code-focused instruction data within the \texttt{Nemotron-SFT-Code} subset. The resulting training stream contains approximately 70M sequences and $\sim$50B tokens.
To study data efficiency, we also consider a \emph{tiny} subset constructed by uniformly subsampling to $\sim$0.8B tokens. For fixed-compute comparisons, we sample with replacement so that the number of optimization steps is matched. In both the full and tiny settings, training examples are sampled with replacement, so the optimization budget is controlled by steps (and thus total tokens processed) rather than epochs.
The tiny-data comparison in \Cref{sec:results_tiny} uses this matched-step construction.

\paragraph{Representation alignment.}
In addition to the denoising loss, we apply layer-wise representation alignment to anchor the student to the frozen AR teacher. The teacher consumes the \emph{clean} sequence $x$ (causal attention), while the student consumes the corrupted $\tilde{x}$ (bidirectional attention). We align the output of every transformer block at every token position using a cosine distance loss, with teacher features stop-gradiented and the teacher run in evaluation mode. We use $\lambda=10$ as the default alignment weight, selected from the ablation in \Cref{tab:repr-ablation}. We additionally enable the PAPL loss~\citep{peng2026planner} in \emph{all} runs with weight $1$. Unless stated otherwise, we treat PAPL as part of the default DLM training recipe and include it in \emph{all} methods (baseline and aligned) so comparisons isolate the effect of representation alignment.
\Cref{sec:ablations} varies the metric, layer set, and $\lambda$ in this alignment term.
\paragraph{Optimization, batching, and evaluation.}
We optimize using AdamW with learning rate $3\times 10^{-4}$, cosine decay, warmup ratio 0.001, weight decay 0.01, gradient clipping (max norm 1.0), and mixed-precision training. The maximum sequence length is 4096, and the global batch size is 96 sequences per optimization step.
We evaluate code generation on HumanEval~\citep{chen2021evaluating}, MBPP~\citep{austin2021program}, and their EvalPlus variants~\citep{liu2023evalplus}. HumanEval uses the canonical docstring prompts, and all evaluations are zero-shot. For decoding, we use P2-self sampling~\citep{peng2025pathplanningmaskeddiffusion} with 128 sampling steps, $\texttt{max\_new\_tokens}=128$, temperature 0.8, and top-$p$ 0.95. Full data preprocessing, optimization, decoding, and evaluation configurations are reported in \Cref{app:data_preprocessing,app:optimization,app:evaluation_details}.

\subsection{Results}
\label{sec:results}

The experiments are organized around four questions induced by \Cref{sec:method}. First, does the alignment term in \Cref{eq:align_loss} improve AR$\rightarrow$DLM adaptation under a matched training budget? Second, do the gains scale with model size? Third, does representation preservation reduce the need to update all parameters? Fourth, does conversion remain effective when the adaptation data is sharply reduced? The following paragraphs answer these questions, respectively.

\begin{figure}[t]
    \centering
    \includegraphics[width=0.42\textwidth]{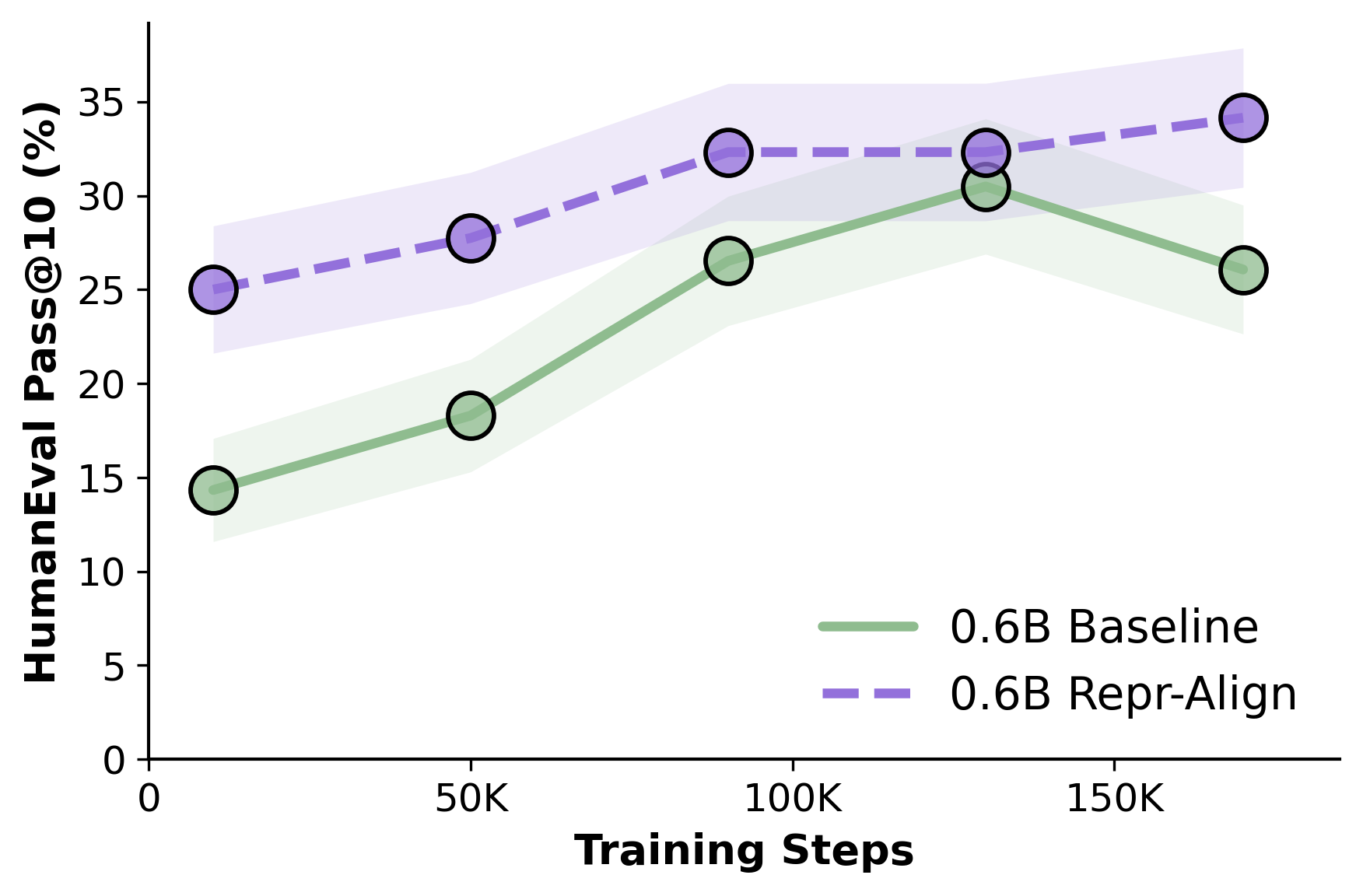}
    \hfill
    \includegraphics[width=0.42\textwidth]{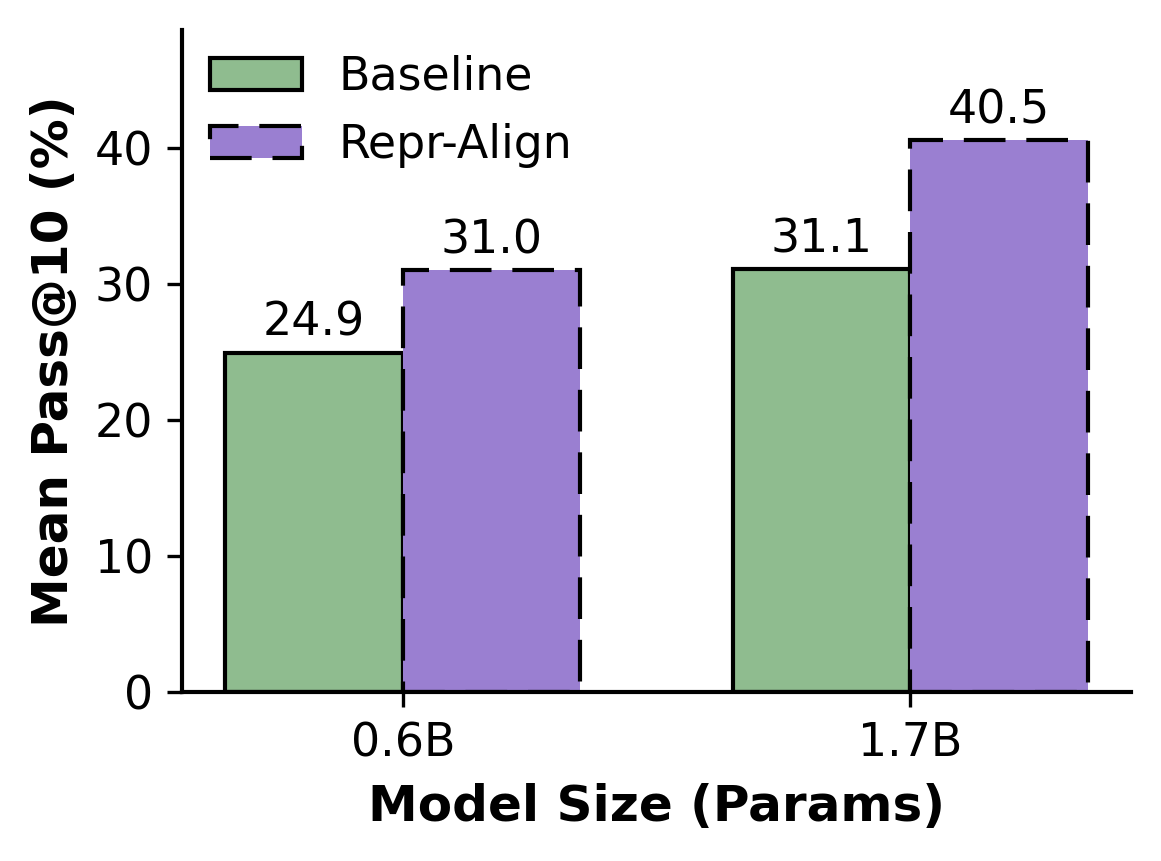}
    \caption{\nameshort improves both adaptation speed and final quality.
\textbf{Left:} HumanEval pass@10 vs. training steps for Qwen3-0.6B during AR$\rightarrow$DLM conversion; adding representation alignment to the frozen AR teacher improves sample efficiency throughout training.
\textbf{Right:} pass@10 results for 0.6B and 1.7B models; representation alignment provides larger gains at 1.7B than at 0.6B.}
    \label{fig:repralign_curve}
\end{figure}

\paragraph{\nameshort improves adaptation efficiency and quality.}
\label{sec:results_alignment}
\Cref{fig:repralign_curve} evaluates autoregressive-to-diffusion conversion on HumanEval under a fixed 200k-step budget.
Starting from the same pretrained Qwen3 checkpoint, the baseline performs masked diffusion training (MDLM-style masking with the shift convention and PAPL), while our method additionally anchors the student to the frozen AR teacher via layer-wise cosine alignment.
\nameshort consistently improves pass@10 throughout training, indicating substantially better sample efficiency during adaptation.
At the 200k-step cutoff, the aligned model achieves higher pass@10 than the baseline (a gain of 6.1 points at 0.6B and 9.4 points at 1.7B), demonstrating that anchoring pretrained representations is beneficial even when the student is trained with a bidirectional denoising objective.
This validates the alignment hypothesis encoded by \Cref{eq:align_loss} under the matched conversion setup in \Cref{sec:method_models}.

\paragraph{Alignment gains grow with model size.}
\label{sec:results_scaling}
\Cref{fig:repralign_curve} summarizes the same conversion procedure across model scales.
For each size, we compare the last checkpoint after 200k steps for the baseline and aligned variants under identical evaluation.
We observe that alignment benefits increase with model capacity: the absolute improvement from alignment is larger for the 1.7B model than for the 0.6B model.
This scaling trend supports the view that AR pretraining learns a strong representational geometry that is increasingly valuable to preserve as model capacity grows, while the remaining adaptation primarily concerns the generation mechanism induced by the attention mask and denoising dynamics.
This directly tests whether the preservation term in \Cref{eq:full_loss} becomes more useful as the AR teacher's representation capacity increases.

\paragraph{oDLM is competitive with public diffusion language models.}
\label{sec:results_public}
\Cref{fig:modelscompare} gives the high-level comparison that motivates the paper: after conversion, oDLM reaches the public frontier of diffusion language models on code-generation pass@10 while using pretrained AR representations rather than retraining a diffusion LM from scratch.
\Cref{tab:code-generation} reports the full benchmark breakdown across HumanEval, HumanEval+, MBPP, and MBPP+ against recent public DLM systems, with code-focused diffusion LMs such as DiffuCoder and Dream-Coder providing closely related context for this evaluation setting~\citep{nie2025llada,dream2025,diffucoder2025,dreamcoder2025}.
The strongest oDLM checkpoint is especially competitive on pass@10, which is the metric most directly tied to iterative diffusion sampling quality.
This comparison evaluates the full objective in \Cref{eq:full_loss} rather than the alignment term in isolation.


\begin{table}[h]
\centering
\caption{Code generation results on HumanEval, HumanEval+, MBPP, and MBPP+ benchmarks, compared with recent public diffusion language models~\citep{nie2025llada,dream2025}.}
\label{tab:code-generation}
\resizebox{\textwidth}{!}{%
\begin{tabular}{lcccccccc}
\toprule
& \multicolumn{2}{c}{\textbf{HumanEval}} & \multicolumn{2}{c}{\textbf{HumanEval+}} & \multicolumn{2}{c}{\textbf{MBPP}} & \multicolumn{2}{c}{\textbf{MBPP+}} \\
\cmidrule(lr){2-3} \cmidrule(lr){4-5} \cmidrule(lr){6-7} \cmidrule(lr){8-9}
\textbf{Model} & Pass@1 & Pass@10 & Pass@1 & Pass@10 & Pass@1 & Pass@10 & Pass@1 & Pass@10 \\
\midrule
LLaDA (8B)        & \underline{35.4}  & 50.0              & \underline{30.5}  & 43.3              & \underline{38.8}  & \underline{53.4}  & \underline{52.6}  & \underline{69.1}  \\
Dream (7B)        & \textbf{56.7}     & \underline{59.2}  & \textbf{50.0}     & \underline{53.7}  & \textbf{55.4}     & \textbf{56.2}     & \textbf{71.5}     & \textbf{72.5}     \\
Mask DFM (1.3B)   & 9.1               & 17.6              & 7.9               & 13.4              & 6.2               & 25.0              & --                & --                \\
Edit Flow (1.3B)  & 12.8              & 24.3              & 10.4              & 20.7              & 10.0              & 36.4              & --                & --                \\
\midrule
oDLM (0.6B)       & 17.87             & 35.98             & 16.40             & 31.10             & 15.70             & 37.00             & 22.99             & 49.47             \\
oDLM (4B)         & 30.49             & \textbf{61.59}    & 26.95             & \textbf{56.10}    & 14.42             & 37.00             & 20.24             & 47.35             \\
\bottomrule
\end{tabular}%
}
\end{table}

\begin{figure}[t]
    \centering
    \includegraphics[width=0.52\textwidth]{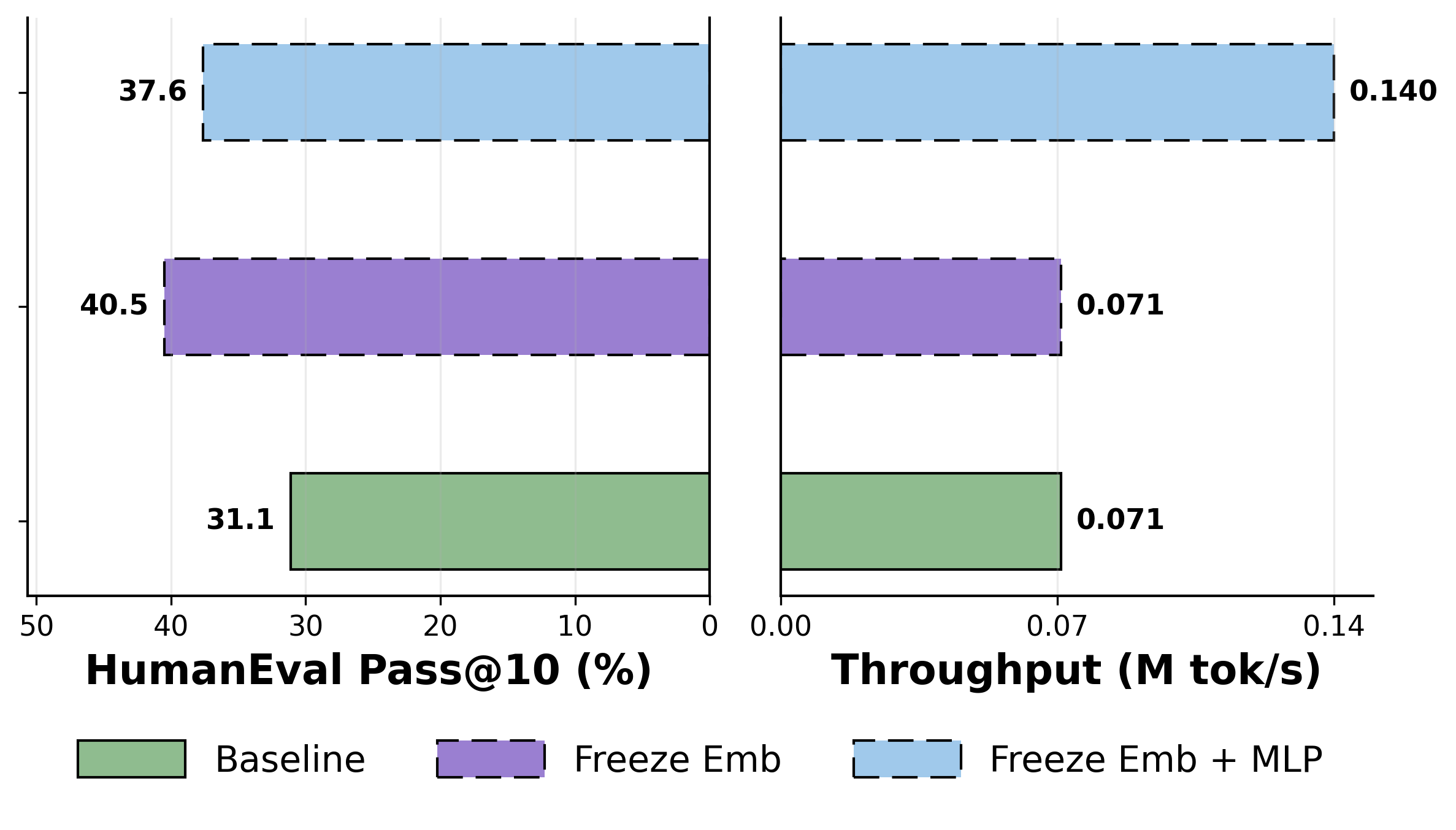}
    \caption{Freezing improves training efficiency with a mild performance gain (1.7B).
All runs use representation alignment and share the same training protocol and budget; we only vary which parameter blocks are frozen.
Freezing embeddings and MLP blocks increases throughput by up to $\sim$2$\times$ while slightly improving HumanEval pass@10.}
    \label{fig:freezing}
\end{figure}

\paragraph{Freezing large blocks improves throughput with a mild quality gain.}
\label{sec:results_freezing}
If AR pretraining already provides strong representations, and representation alignment preserves the teacher's internal feature space during conversion, then AR$\rightarrow$DLM adaptation should not require updating the entire network.
We test this by freezing large parameter blocks in the aligned student while keeping the rest of the training protocol fixed.
\Cref{fig:freezing} shows a favorable performance--efficiency trade-off at the 1.7B scale: freezing token embeddings and, more aggressively, freezing both embeddings and MLP blocks yields a substantial increase in training throughput (up to $\sim$2$\times$), while maintaining and even slightly improving HumanEval pass@10.
This supports the view that, once representations are anchored, the remaining learning signal is concentrated in adapting the denoising mechanism under bidirectional attention, and freezing provides a practical knob to reduce conversion cost without sacrificing quality.
This result tests the mechanism-adaptation interpretation of \Cref{sec:method_alignment}; the freezing protocol is specified in \Cref{app:freezing_protocol}.

\begin{figure}[t]
  \centering
  \begin{subfigure}[t]{0.46\linewidth}
    \centering
    \includegraphics[width=\linewidth]{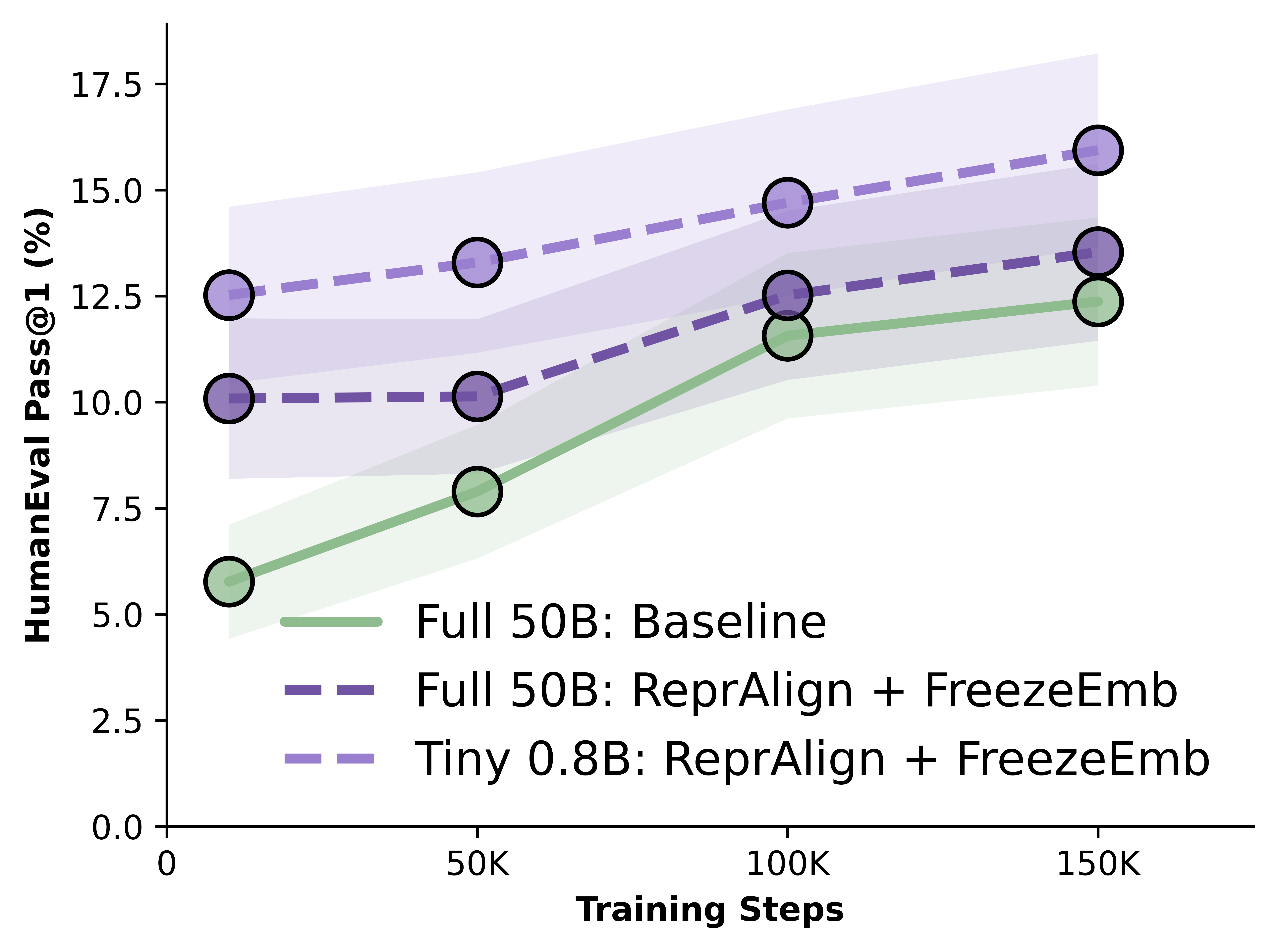}
    \caption{HumanEval pass@1.}
    \label{fig:tiny-pass1}
  \end{subfigure}
  \hfill
  \begin{subfigure}[t]{0.46\linewidth}
    \centering
    \includegraphics[width=\linewidth]{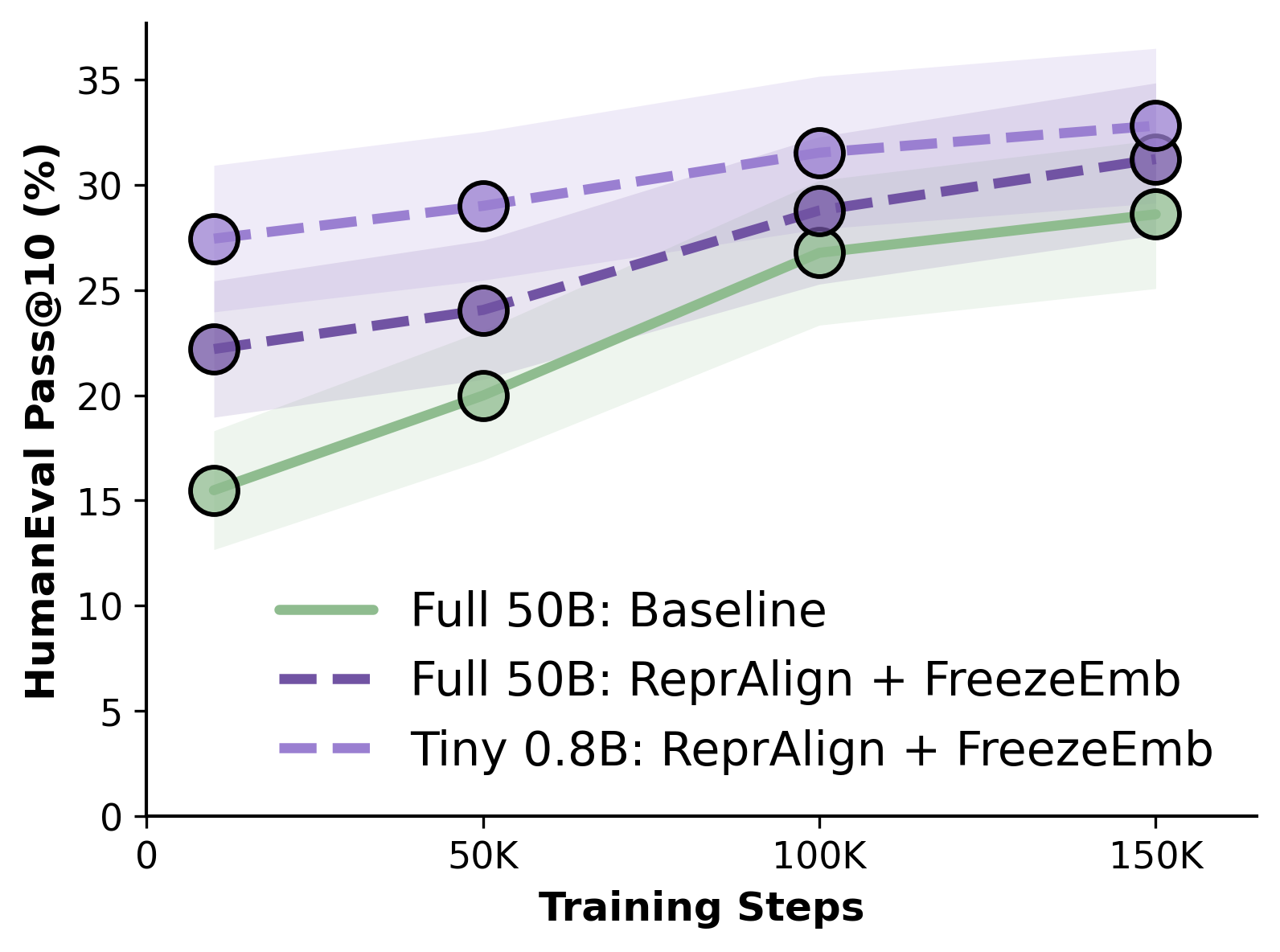}
    \caption{HumanEval pass@10.}
    \label{fig:tiny-pass10}
  \end{subfigure}
  \caption{Alignment is not data-hungry: a tiny subset can improve conversion. The \emph{tiny} run trains on a 0.8B-token random subsample instead of the full 50B-token stream. With representation alignment, the tiny subset yields consistently higher pass@1 (left) and pass@10 (right), supporting the view that AR$\rightarrow$DLM conversion is primarily mechanism adaptation without massive training data.}
  \label{fig:tiny}
\end{figure}

\paragraph{Alignment is especially data-efficient: tiny data can improve conversion.}
\label{sec:results_tiny}
To test whether AR-to-DLM conversion is inherently data-hungry, we reduce the training corpus from the full 50B-token Nemotron-SFT-Code stream to a \emph{tiny} 0.8B-token random subsample, while keeping the optimization budget fixed (same batch size, sequence length, and number of steps; tiny data is sampled with replacement).
\Cref{fig:tiny} shows a clear and somewhat surprising outcome: with representation alignment, training on the tiny subset yields \emph{higher} HumanEval pass@10 than training on the full data stream at the same training steps.
This indicates that, once pretrained AR representations are preserved, the remaining learning problem is primarily \emph{mechanism adaptation} (adapting bidirectional attention and denoising dynamics), which can be accomplished with remarkably little data under a fixed-step budget.
This validates the low-data implication of anchoring the student to the teacher representation in \Cref{eq:align_loss}.

\paragraph{Practical takeaway.}
\label{sec:results_takeaway}
Across scales and data regimes, our experiments support a simple recipe for training diffusion language models efficiently.
Given a pretrained AR transformer, we can convert it into a competitive masked diffusion model by switching to bidirectional attention, training with an MDLM-style denoising objective with the shift convention, and anchoring intermediate representations to the frozen AR teacher with a layer-wise cosine loss.
When compute is constrained, freezing token embeddings and even MLP blocks further improves efficiency with little or no degradation in quality.
When data is constrained, alignment enables effective adaptation even on a tiny subset, suggesting that large diffusion pretraining costs are not intrinsic, but largely reflect relearning representations that AR pretraining already provides.

\subsection{Ablation Studies}
\label{sec:ablations}

We ablate the main design choices in \Cref{eq:align_loss,eq:full_loss}: the distance used to match hidden states, the strength $\lambda$ of the alignment term, and the set of layers included in the alignment loss.
All ablations use the same AR$\rightarrow$DLM conversion protocol unless otherwise stated: the student is initialized from the AR checkpoint, trained with bidirectional masked denoising, evaluated on HumanEval, and decoded with the same sampling configuration.
This setup isolates whether the gain comes from preserving the pretrained representation geometry rather than from changes in architecture or decoding.
\Cref{tab:repr-ablation} summarizes the main ablation outcomes. Cosine alignment is the best default metric; $\lambda=10$ gives the strongest overall anchor; and all-layer alignment provides the best pass@1 while maintaining competitive pass@10.

\begin{table*}[t]
    \centering
    \scriptsize
    \setlength{\tabcolsep}{6pt}
    \caption{
    Ablations of representation alignment on HumanEval.
    }
    \label{tab:repr-ablation}
    \begin{tabular*}{0.9\textwidth}{@{\extracolsep{\fill}}lcc lcc lcc@{}}
    \toprule
    \multicolumn{3}{c}{\textbf{Metric}} &
    \multicolumn{3}{c}{\textbf{Weight}} &
    \multicolumn{3}{c}{\textbf{Layers}} \\
    \cmidrule(lr){1-3}\cmidrule(lr){4-6}\cmidrule(lr){7-9}
    \textbf{Variant} & \textbf{pass@1} & \textbf{pass@10} &
    \textbf{$\lambda$} & \textbf{pass@1} & \textbf{pass@10} &
    \textbf{Variant} & \textbf{pass@1} & \textbf{pass@10} \\
    \midrule
    L2     & 12.00 & 25.00 & 1  & 7.50  & 22.60 & Lower  & 11.28 & 25.00 \\
    Cosine & \textbf{18.00} & \textbf{31.00} & 5  & 9.90  & 26.80 & Middle & 15.73 & 31.10 \\
           &       &       & 10 & \textbf{18.00} & \textbf{31.00} & Upper  & 8.96  & \textbf{31.71} \\
           &       &       & 20 & 11.30 & 29.30 & All    & \textbf{18.00} & 31.00 \\
    \bottomrule
    \end{tabular*}
    \end{table*}
\paragraph{Cosine alignment is better than matching hidden states by L2.}
\label{sec:ablation_metric}
We first compare two natural choices for the representation loss: an L2 loss on hidden states and a cosine-distance loss. 
This directly tests the cosine-distance choice in \Cref{eq:align_loss}.
Cosine alignment improves HumanEval pass@1 from $12.0$ to $18.0$ and pass@10 from $25.0$ to $31.0$, suggesting that the useful signal in the AR teacher is primarily geometric rather than metric.
Because transformer hidden-state norms vary across layers, tokens, and training states, an L2 objective can be dominated by scale; cosine alignment instead preserves representation direction.
We therefore use cosine alignment as the default representation loss.

\paragraph{The alignment weight controls a trade-off between anchoring and adaptation.}
\label{sec:ablation_weight}
We next vary the alignment weight $\lambda$ in \Cref{eq:full_loss}, which controls the trade-off between anchoring to the AR teacher and adapting to the denoising objective.
The sweep shows a clear trade-off: weak alignment underuses the AR teacher, while excessive anchoring constrains the diffusion student.
Performance rises from $\lambda=1$ to $\lambda=10$, reaching $18.0$ pass@1 and $31.0$ pass@10, but drops at $\lambda=20$.
This supports treating representation alignment as an auxiliary constraint rather than exact teacher imitation.
We therefore set $\lambda=10$ by default.

\paragraph{Where to align matters.}
\label{sec:ablation_layers}
Finally, we partition hidden states into lower, middle, and upper thirds to test where the AR teacher is most useful.
Middle- and upper-layer anchors improve pass@10, but only all-layer alignment gives the strongest pass@1 while preserving competitive pass@10.
The transferable signal therefore appears distributed across depth.
We therefore use all-layer cosine alignment as the default.

\section{Conclusion}
\label{sec:conclusion}
In this work, we challenge the prevailing dichotomy between autoregressive and diffusion language models. Our empirical investigation substantiates that the rich semantic representations acquired during standard autoregressive pretraining are not specific to sequential decoding, but are broadly applicable to non-autoregressive generation. By introducing \nameshort, we demonstrate that it is possible to inherit this semantic topology directly. We achieve state-of-the-art performance with a fraction of the standard pretraining cost.

Our findings have broader implications for discrete generation. Alternative frameworks such as uniform, latent, and simplex diffusion models have historically struggled to scale, often due to optimization difficulties. By anchoring new generation mechanisms to a robust AR prior, \nameshort offers a recipe for adapting pretrained representations rather than training each paradigm ab initio. We hope this encourages a shift toward mechanism alignment as a practical way to unlock new forms of generation.

\begin{ack}
We thank Jarrid Rector-brooks for helpful discussions and assistance. F.Z.P. and A.R.Z. are partially supported by NIH R01HL169347.
\end{ack}

\bibliographystyle{plainnat}
\bibliography{references}

@inproceedings{gong2025scalingdiffusionlanguagemodels,
  title     = {Scaling Diffusion Language Models via Adaptation from Autoregressive Models},
  author    = {Shansan Gong and Shivam Agarwal and Yizhe Zhang and Jiacheng Ye and Lin Zheng and Mukai Li and Chenxin An and Peilin Zhao and Wei Bi and Jiawei Han and Hao Peng and Lingpeng Kong},
  booktitle = {International Conference on Learning Representations},
  year      = {2025}
}

@inproceedings{yu2025repa,
  title     = {Representation Alignment for Generation: Training Diffusion Transformers Is Easier Than You Think},
  author    = {Sihyun Yu and Sangkyung Kwak and Huiwon Jang and Jongheon Jeong and Jonathan Huang and Jinwoo Shin and Saining Xie},
  booktitle = {International Conference on Learning Representations},
  year      = {2025}
}

@article{singh2025irepa,
  title   = {What Matters for Representation Alignment: Global Information or Spatial Structure?},
  author  = {Singh, Jaskirat and Leng, Xingjian and Wu, Zongze and Zheng, Liang and Zhang, Richard and Shechtman, Eli and Xie, Saining},
  journal = {arXiv preprint arXiv:2512.10794},
  year    = {2025}
}

@article{wu2025representation,
  title   = {Representation Entanglement for Generation: Training Diffusion Transformers Is Much Easier Than You Think},
  author  = {Wu, Ge and Zhang, Shen and Shi, Ruijing and Gao, Shanghua and Chen, Zhenyuan and Wang, Lei and Chen, Zhaowei and Gao, Hongcheng and Tang, Yao and Yang, Jian and Cheng, Ming-Ming and Li, Xiang},
  journal = {arXiv preprint arXiv:2507.01467},
  year    = {2025}
}

@article{jiang2025sra,
  title   = {No Other Representation Component Is Needed: Diffusion Transformers Can Provide Representation Guidance by Themselves},
  author  = {Jiang, Dengyang and Wang, Mengmeng and Li, Liuzhuozheng and Zhang, Lei and Wang, Haoyu and Wei, Wei and Dai, Guang and Zhang, Yanning and Wang, Jingdong},
  journal = {arXiv preprint arXiv:2505.02831},
  year    = {2025}
}

@article{nie2024scalingmaskeddiffusionmodels,
  title   = {Scaling up Masked Diffusion Models on Text},
  author  = {Shen Nie and Fengqi Zhu and Chao Du and Tianyu Pang and Qian Liu and Guangtao Zeng and Min Lin and Chongxuan Li},
  year    = {2025},
  journal = {International Conference on Learning Representations}
}

@InProceedings{Chang_2022_CVPR,
  author    = {Chang, Huiwen and Zhang, Han and Jiang, Lu and Liu, Ce and Freeman, William T.},
  title     = {MaskGIT: Masked Generative Image Transformer},
  booktitle = {Proceedings of the IEEE/CVF Conference on Computer Vision and Pattern Recognition (CVPR)},
  year      = {2022},
  pages     = {11315-11325}
}

@article{2,
  title  = {DPLM-2: A Multimodal Diffusion Protein Language Model},
  author = {Xinyou Wang and Zaixiang Zheng and Fei Ye and Dongyu Xue and Shujian Huang and Quanquan Gu},
  journal= {ArXiv},
  year   = {2024},
  volume = {abs/2410.13782},
  url    = {https://api.semanticscholar.org/CorpusID:273403705}
}

@inproceedings{mdlm,
  title     = {Simple and Effective Masked Diffusion Language Models},
  author    = {Subham Sekhar Sahoo and Marianne Arriola and Aaron Gokaslan and Edgar Mariano Marroquin and Alexander M Rush and Yair Schiff and Justin T Chiu and Volodymyr Kuleshov},
  booktitle = {The Thirty-eighth Annual Conference on Neural Information Processing Systems},
  year      = {2024}
}

@inproceedings{Lou2023DiscreteDM,
  title     = {Discrete Diffusion Modeling by Estimating the Ratios of the Data Distribution},
  author    = {Aaron Lou and Chenlin Meng and Stefano Ermon},
  booktitle = {International Conference on Machine Learning},
  year      = {2023}
}

@misc{peng2025pathplanningmaskeddiffusion,
      title={Path Planning for Masked Diffusion Model Sampling}, 
      author={Fred Zhangzhi Peng and Zachary Bezemek and Sawan Patel and Jarrid Rector-Brooks and Sherwood Yao and Avishek Joey Bose and Alexander Tong and Pranam Chatterjee},
      year={2025},
      eprint={2502.03540},
      archivePrefix={arXiv},
      primaryClass={cs.LG},
      url={https://arxiv.org/abs/2502.03540}, 
}

@article{Gulrajani2023LikelihoodBasedDL,
  title   = {Likelihood-Based Diffusion Language Models},
  author  = {Ishaan Gulrajani and Tatsunori Hashimoto},
  journal = {Neural Information Processing Systems},
  year    = {2023}
}

@article{Touvron2023Llama2O,
  title   = {Llama 2: Open Foundation and Fine-Tuned Chat Models},
  author  = {Hugo Touvron and Louis Martin and Kevin R. Stone and Peter Albert and Amjad Almahairi and Yasmine Babaei and Nikolay Bashlykov and Soumya Batra and Prajjwal Bhargava and Shruti Bhosale and Daniel M. Bikel and Lukas Blecher and Cristian Cant{\'o}n Ferrer and Moya Chen and Guillem Cucurull and David Esiobu and Jude Fernandes and Jeremy Fu and Wenyin Fu and Brian Fuller and Cynthia Gao and Vedanuj Goswami and Naman Goyal and Anthony S. Hartshorn and Saghar Hosseini and Rui Hou and Hakan Inan and Marcin Kardas and Viktor Kerkez and Madian Khabsa and Isabel M. Kloumann and A. V. Korenev and Punit Singh Koura and Marie-Anne Lachaux and Thibaut Lavril and Jenya Lee and Diana Liskovich and Yinghai Lu and Yuning Mao and Xavier Martinet and Todor Mihaylov and Pushkar Mishra and Igor Molybog and Yixin Nie and Andrew Poulton and Jeremy Reizenstein and Rashi Rungta and Jacob Hilton and Reiichiro Nakano and Christopher Hesse and John Schulman and Rashi Rungta and Kalyan Saladi and Alan Schelten and Ruan Silva and Eric Michael Smith and R. Subramanian and Xia Tan and Binh Tang and Ross Taylor and Adina Williams and Jian Xiang Kuan and Puxin Xu and Zhengxu Yan and Iliyan Zarov and Yuchen Zhang and Angela Fan and Melissa Hall Melanie Kambadur and Sharan Narang and Aur{\'e}lien Rodriguez and Robert Stojnic and Sergey Edunov and Thomas Scialom},
  journal = {arXiv},
  year    = {2023}
}

@article{radford2019language,
  title   = {Language Models are Unsupervised Multitask Learners},
  author  = {Radford, Alec and Wu, Jeff and Child, Rewon and Luan, David and Amodei, Dario and Sutskever, Ilya},
  year    = {2019},
  journal = {preprint}
}

@article{Austin2021StructuredDD,
  author  = {Jacob Austin and Daniel D. Johnson and Jonathan Ho and Daniel Tarlow and Rianne van den Berg},
  title   = {Structured Denoising Diffusion Models in Discrete State-Spaces},
  journal = {arXiv},
  year    = {2021}
}

@misc{campbell2022continuoustimeframeworkdiscrete,
  title   = {A Continuous Time Framework for Discrete Denoising Models},
  author  = {Andrew Campbell and Joe Benton and Valentin De Bortoli and Tom Rainforth and George Deligiannidis and Arnaud Doucet},
  year    = {2022},
  journal = {Neural Information Processing Systems}
}

@article{shih2022traininginferenceanyorderautoregressive,
  title   = {Training and Inference on Any-Order Autoregressive Models the Right Way},
  author  = {Andy Shih and Dorsa Sadigh and Stefano Ermon},
  year    = {2022},
  journal = {Neural Information Processing Systems}
}

@article{ho2020denoising,
  title   = {Denoising diffusion probabilistic models},
  author  = {Ho, Jonathan and Jain, Ajay and Abbeel, Pieter},
  journal = {Advances in neural information processing systems},
  volume  = {33},
  pages   = {6840--6851},
  year    = {2020}
}

@article{achiam2023gpt,
  title   = {Gpt-4 technical report},
  author  = {Achiam, Josh and Adler, Steven and Agarwal, Sandhini and Ahmad, Lama and Akkaya, Ilge and Aleman, Florencia Leoni and Almeida, Diogo and Altenschmidt, Janko and Altman, Sam and Anadkat, Shyamal and others},
  journal = {arXiv},
  year    = {2023}
}

@InProceedings{pmlr-v37-sohl-dickstein15,
  title     = {Deep Unsupervised Learning using Nonequilibrium Thermodynamics},
  author    = {Sohl-Dickstein, Jascha and Weiss, Eric and Maheswaranathan, Niru and Ganguli, Surya},
  booktitle = {Proceedings of the 32nd International Conference on Machine Learning},
  pages     = {2256--2265},
  year      = {2015},
  editor    = {Bach, Francis and Blei, David},
  volume    = {37},
  series    = {Proceedings of Machine Learning Research},
  address   = {Lille, France},
  month     = {07--09 Jul},
  publisher = {PMLR},
  url       = {https://proceedings.mlr.press/v37/sohl-dickstein15.html}
}

@inproceedings{
peng2026planner,
title={Planner Aware Path Learning in Diffusion Language Models Training},
author={Fred Zhangzhi Peng and Zachary Bezemek and Jarrid Rector-Brooks and Shuibai Zhang and Michael M. Bronstein and Anru Zhang and Joey Bose and Alexander Tong},
booktitle={The Fourteenth International Conference on Learning Representations},
year={2026},
url={https://openreview.net/forum?id=lAlI5FuIf7}
}

@misc{nvidia2025nvidianemotronnano2,
      title={NVIDIA Nemotron Nano 2: An Accurate and Efficient Hybrid Mamba-Transformer Reasoning Model}, 
      author={NVIDIA and : and Aarti Basant and Abhijit Khairnar and Abhijit Paithankar and Abhinav Khattar and Adithya Renduchintala and Aditya Malte and Akhiad Bercovich and Akshay Hazare and Alejandra Rico and Aleksander Ficek and Alex Kondratenko and Alex Shaposhnikov and Alexander Bukharin and Ali Taghibakhshi and Amelia Barton and Ameya Sunil Mahabaleshwarkar and Amy Shen and Andrew Tao and Ann Guan and Anna Shors and Anubhav Mandarwal and Arham Mehta and Arun Venkatesan and Ashton Sharabiani and Ashwath Aithal and Ashwin Poojary and Ayush Dattagupta and Balaram Buddharaju and Banghua Zhu and Barnaby Simkin and Bilal Kartal and Bita Darvish Rouhani and Bobby Chen and Boris Ginsburg and Brandon Norick and Brian Yu and Bryan Catanzaro and Charles Wang and Charlie Truong and Chetan Mungekar and Chintan Patel and Chris Alexiuk and Christian Munley and Christopher Parisien and Dan Su and Daniel Afrimi and Daniel Korzekwa and Daniel Rohrer and Daria Gitman and David Mosallanezhad and Deepak Narayanan and Dima Rekesh and Dina Yared and Dmytro Pykhtar and Dong Ahn and Duncan Riach and Eileen Long and Elliott Ning and Eric Chung and Erick Galinkin and Evelina Bakhturina and Gargi Prasad and Gerald Shen and Haifeng Qian and Haim Elisha and Harsh Sharma and Hayley Ross and Helen Ngo and Herman Sahota and Hexin Wang and Hoo Chang Shin and Hua Huang and Iain Cunningham and Igor Gitman and Ivan Moshkov and Jaehun Jung and Jan Kautz and Jane Polak Scowcroft and Jared Casper and Jian Zhang and Jiaqi Zeng and Jimmy Zhang and Jinze Xue and Jocelyn Huang and Joey Conway and John Kamalu and Jonathan Cohen and Joseph Jennings and Julien Veron Vialard and Junkeun Yi and Jupinder Parmar and Kari Briski and Katherine Cheung and Katherine Luna and Keith Wyss and Keshav Santhanam and Kezhi Kong and Krzysztof Pawelec and Kumar Anik and Kunlun Li and Kushan Ahmadian and Lawrence McAfee and Laya Sleiman and Leon Derczynski and Luis Vega and Maer Rodrigues de Melo and Makesh Narsimhan Sreedhar and Marcin Chochowski and Mark Cai and Markus Kliegl and Marta Stepniewska-Dziubinska and Matvei Novikov and Mehrzad Samadi and Meredith Price and Meriem Boubdir and Michael Boone and Michael Evans and Michal Bien and Michal Zawalski and Miguel Martinez and Mike Chrzanowski and Mohammad Shoeybi and Mostofa Patwary and Namit Dhameja and Nave Assaf and Negar Habibi and Nidhi Bhatia and Nikki Pope and Nima Tajbakhsh and Nirmal Kumar Juluru and Oleg Rybakov and Oleksii Hrinchuk and Oleksii Kuchaiev and Oluwatobi Olabiyi and Pablo Ribalta and Padmavathy Subramanian and Parth Chadha and Pavlo Molchanov and Peter Dykas and Peter Jin and Piotr Bialecki and Piotr Januszewski and Pradeep Thalasta and Prashant Gaikwad and Prasoon Varshney and Pritam Gundecha and Przemek Tredak and Rabeeh Karimi Mahabadi and Rajen Patel and Ran El-Yaniv and Ranjit Rajan and Ria Cheruvu and Rima Shahbazyan and Ritika Borkar and Ritu Gala and Roger Waleffe and Ruoxi Zhang and Russell J. Hewett and Ryan Prenger and Sahil Jain and Samuel Kriman and Sanjeev Satheesh and Saori Kaji and Sarah Yurick and Saurav Muralidharan and Sean Narenthiran and Seonmyeong Bak and Sepehr Sameni and Seungju Han and Shanmugam Ramasamy and Shaona Ghosh and Sharath Turuvekere Sreenivas and Shelby Thomas and Shizhe Diao and Shreya Gopal and Shrimai Prabhumoye and Shubham Toshniwal and Shuoyang Ding and Siddharth Singh and Siddhartha Jain and Somshubra Majumdar and Soumye Singhal and Stefania Alborghetti and Syeda Nahida Akter and Terry Kong and Tim Moon and Tomasz Hliwiak and Tomer Asida and Tony Wang and Tugrul Konuk and Twinkle Vashishth and Tyler Poon and Udi Karpas and Vahid Noroozi and Venkat Srinivasan and Vijay Korthikanti and Vikram Fugro and Vineeth Kalluru and Vitaly Kurin and Vitaly Lavrukhin and Wasi Uddin Ahmad and Wei Du and Wonmin Byeon and Ximing Lu and Xin Dong and Yashaswi Karnati and Yejin Choi and Yian Zhang and Ying Lin and Yonggan Fu and Yoshi Suhara and Zhen Dong and Zhiyu Li and Zhongbo Zhu and Zijia Chen},
      year={2025},
      eprint={2508.14444},
      archivePrefix={arXiv},
      primaryClass={cs.CL},
      url={https://arxiv.org/abs/2508.14444}, 
}

@misc{dream2025,
  title        = {Dream 7B: Diffusion Large Language Models},
  author       = {Jiacheng Ye and Zhihui Xie and Lin Zheng and Jiahui Gao and Zirui Wu and Xin Jiang and Zhenguo Li and Lingpeng Kong},
  year         = {2025},
  eprint       = {2508.15487},
  archivePrefix= {arXiv},
  primaryClass = {cs.CL},
  url          = {https://arxiv.org/abs/2508.15487}
}

@misc{dreamcoder2025,
  title        = {Dream-Coder 7B: An Open Diffusion Language Model for Code},
  author       = {Zhihui Xie and Jiacheng Ye and Lin Zheng and Jiahui Gao and Jingwei Dong and Zirui Wu and Xueliang Zhao and Shansan Gong and Xin Jiang and Zhenguo Li and Lingpeng Kong},
  year         = {2025},
  eprint       = {2509.01142},
  archivePrefix= {arXiv},
  primaryClass = {cs.CL},
  url          = {https://arxiv.org/abs/2509.01142}
}

@misc{efficientdlm2025,
  title        = {Efficient-DLM: From Autoregressive to Diffusion Language Models, and Beyond in Speed},
  author       = {Yonggan Fu and Lexington Whalen and Zhifan Ye and Xin Dong and Shizhe Diao and Jingyu Liu and Chengyue Wu and Hao Zhang and Enze Xie and Song Han and Maksim Khadkevich and Jan Kautz and Yingyan Celine Lin and Pavlo Molchanov},
  year         = {2025},
  eprint       = {2512.14067},
  archivePrefix= {arXiv},
  primaryClass = {cs.CL},
  url          = {https://arxiv.org/abs/2512.14067}
}

@misc{xue2025anyordergpt,
  title        = {Any-Order GPT as Masked Diffusion Model: Decoupling Formulation and Architecture},
  author       = {Shuchen Xue and Tianyu Xie and Tianyang Hu and Zijin Feng and Jiacheng Sun and Kenji Kawaguchi and Zhenguo Li and Zhi-Ming Ma},
  year         = {2025},
  eprint       = {2506.19935},
  archivePrefix= {arXiv},
  primaryClass = {cs.LG},
  url          = {https://arxiv.org/abs/2506.19935}
}

@misc{diffucoder2025,
  title        = {DiffuCoder: Understanding and Improving Masked Diffusion Models for Code Generation},
  author       = {Shansan Gong and Ruixiang Zhang and Huangjie Zheng and Jiatao Gu and Navdeep Jaitly and Lingpeng Kong and Yizhe Zhang},
  year         = {2025},
  eprint       = {2506.20639},
  archivePrefix= {arXiv},
  primaryClass = {cs.CL},
  url          = {https://arxiv.org/abs/2506.20639}
}

@misc{nie2025llada,
  title        = {Large Language Diffusion Models},
  author       = {Shen Nie and Fengqi Zhu and Zebin You and Xiaolu Zhang and Jingyang Ou and Jun Hu and Jun Zhou and Yankai Lin and Ji-Rong Wen and Chongxuan Li},
  year         = {2025},
  eprint       = {2502.09992},
  archivePrefix= {arXiv},
  primaryClass = {cs.CL},
  url          = {https://arxiv.org/abs/2502.09992}
}

@inproceedings{li2022diffusionlm,
  title     = {Diffusion-LM Improves Controllable Text Generation},
  author    = {Xiang Lisa Li and John Thickstun and Ishaan Gulrajani and Percy Liang and Tatsunori B. Hashimoto},
  booktitle = {Advances in Neural Information Processing Systems},
  year      = {2022}
}

@inproceedings{yang2019xlnet,
  title     = {{XLNet}: Generalized Autoregressive Pretraining for Language Understanding},
  author    = {Zhilin Yang and Zihang Dai and Yiming Yang and Jaime Carbonell and Ruslan Salakhutdinov and Quoc V. Le},
  booktitle = {Advances in Neural Information Processing Systems},
  year      = {2019}
}

@inproceedings{ghazvininejad2019maskpredict,
  title     = {Mask-Predict: Parallel Decoding of Conditional Masked Language Models},
  author    = {Marjan Ghazvininejad and Omer Levy and Yinhan Liu and Luke Zettlemoyer},
  booktitle = {Proceedings of the 2019 Conference on Empirical Methods in Natural Language Processing and the 9th International Joint Conference on Natural Language Processing},
  pages     = {6112--6121},
  year      = {2019},
  publisher = {Association for Computational Linguistics},
  doi       = {10.18653/v1/D19-1633},
  url       = {https://aclanthology.org/D19-1633/}
}

@misc{qwen3technicalreport,
  title        = {Qwen3 Technical Report},
  author       = {An Yang and Anfeng Li and Baosong Yang and Beichen Zhang and Binyuan Hui and Bo Zheng and Bowen Yu and Chang Gao and Chengen Huang and Chenxu Lv and Chujie Zheng and Dayiheng Liu and Fan Zhou and Fei Huang and Feng Hu and Hao Ge and Haoran Wei and Huan Lin and Jialong Tang and Jian Yang and Jianhong Tu and Jianwei Zhang and Jianxin Yang and Jiaxi Yang and Jing Zhou and Jingren Zhou and Junyang Lin and Kai Dang and Keqin Bao and Kexin Yang and Le Yu and Lianghao Deng and Mei Li and Mingfeng Xue and Mingze Li and Pei Zhang and Peng Wang and Qin Zhu and Rui Men and Ruize Gao and Shixuan Liu and Shuang Luo and Tianhao Li and Tianyi Tang and Wenbiao Yin and Xingzhang Ren and Xinyu Wang and Xinyu Zhang and Xuancheng Ren and Yang Fan and Yang Su and Yichang Zhang and Yinger Zhang and Yu Wan and Yuqiong Liu and Zekun Wang and Zeyu Cui and Zhenru Zhang and Zhipeng Zhou and Zihan Qiu},
  year         = {2025},
  eprint       = {2505.09388},
  archivePrefix= {arXiv},
  primaryClass = {cs.CL},
  url          = {https://arxiv.org/abs/2505.09388}
}

@misc{chen2021evaluating,
  title        = {Evaluating Large Language Models Trained on Code},
  author       = {Mark Chen and Jerry Tworek and Heewoo Jun and Qiming Yuan and Henrique Ponde de Oliveira Pinto and Jared Kaplan and Harri Edwards and Yuri Burda and Nicholas Joseph and Greg Brockman and Alex Ray and Raul Puri and Gretchen Krueger and Michael Petrov and Heidy Khlaaf and Girish Sastry and Pamela Mishkin and Brooke Chan and Scott Gray and Nick Ryder and Mikhail Pavlov and Alethea Power and Lukasz Kaiser and Mohammad Bavarian and Clemens Winter and Philippe Tillet and Felipe Petroski Such and Dave Cummings and Matthias Plappert and Fotios Chantzis and Elizabeth Barnes and Ariel Herbert-Voss and William Hebgen Guss and Alex Nichol and Alex Paino and Nikolas Tezak and Jie Tang and Igor Babuschkin and Suchir Balaji and Shantanu Jain and William Saunders and Christopher Hesse and Andrew N. Carr and Jan Leike and Josh Achiam and Vedant Misra and Evan Morikawa and Alec Radford and Matthew Knight and Miles Brundage and Mira Murati and Katie Mayer and Peter Welinder and Bob McGrew and Dario Amodei and Sam McCandlish and Ilya Sutskever and Wojciech Zaremba},
  year         = {2021},
  eprint       = {2107.03374},
  archivePrefix= {arXiv},
  primaryClass = {cs.LG},
  url          = {https://arxiv.org/abs/2107.03374}
}

@misc{austin2021program,
  title        = {Program Synthesis with Large Language Models},
  author       = {Jacob Austin and Augustus Odena and Maxwell Nye and Maarten Bosma and Henryk Michalewski and David Dohan and Ellen Jiang and Carrie Cai and Michael Terry and Quoc Le and Charles Sutton},
  year         = {2021},
  eprint       = {2108.07732},
  archivePrefix= {arXiv},
  primaryClass = {cs.PL},
  url          = {https://arxiv.org/abs/2108.07732}
}

@inproceedings{liu2023evalplus,
  title     = {Is Your Code Generated by ChatGPT Really Correct? Rigorous Evaluation of Large Language Models for Code Generation},
  author    = {Jiawei Liu and Chunqiu Steven Xia and Yuyao Wang and Lingming Zhang},
  booktitle = {Advances in Neural Information Processing Systems},
  year      = {2023},
  url       = {https://openreview.net/forum?id=1qvx610Cu7}
}

\newpage
\appendix
\makeatletter
\section*{Appendix Contents}
\begingroup
\small
\setcounter{tocdepth}{2}
\@starttoc{apc}
\endgroup
\vspace{0.75em}
\let\appendixsection\section
\let\appendixsubsection\subsection
\renewcommand{\section}[1]{%
  \appendixsection{#1}%
  \addcontentsline{apc}{section}{\protect\numberline{\thesection}#1}%
}
\renewcommand{\subsection}[1]{%
  \appendixsubsection{#1}%
  \addcontentsline{apc}{subsection}{\protect\numberline{\thesubsection}#1}%
}
\makeatother

\section{Extended Related Work}
\label{app:extended_related_work}

This appendix provides a more detailed comparison with the closest lines of work. 
The purpose is not to survey all diffusion or language-generation methods exhaustively, but to clarify the specific position of our method: we study AR$\rightarrow$DLM conversion through \emph{representation preservation}. 
Existing conversion methods show that pretrained autoregressive checkpoints are useful initializations for diffusion language models, but they primarily adapt the objective, attention pattern, masking convention, or sampling procedure. 
Our method instead explicitly preserves the hidden-state geometry of the pretrained AR model by anchoring the DLM student to a frozen same-architecture AR teacher.
This appendix supports the positioning claims made in \Cref{sec:related_work} and the methodological distinction formalized in \Cref{sec:method_alignment}.

\subsection{Diffusion language models}
\label{app:rw_dlm}

Diffusion language models have been developed through several distinct formulations. 
Early approaches introduced diffusion into language generation by applying continuous diffusion over word embeddings or latent representations~\citep{li2022diffusionlm}. 
In parallel, discrete diffusion models define corruption and denoising processes directly over categorical state spaces, including structured discrete transition kernels, continuous-time Markov formulations, ratio-estimation objectives, and likelihood-based language-modeling objectives~\citep{Austin2021StructuredDD,campbell2022continuoustimeframeworkdiscrete,Lou2023DiscreteDM,Gulrajani2023LikelihoodBasedDL}. 
More recent masked diffusion language models simplify the discrete diffusion process by using absorbing mask corruption and masked-token denoising objectives, yielding strong likelihood and generation performance with a training objective closely related to masked language modeling~\citep{mdlm,nie2024scalingmaskeddiffusionmodels}. 
Large-scale systems such as LLaDA and Dream further demonstrate that diffusion language models can support instruction following, reasoning, and general-purpose language generation at billion-parameter scale~\citep{nie2025llada,dream2025}.

Our work is orthogonal to the choice of diffusion parameterization. 
We use a masked denoising objective as the target conversion objective, but we do not propose a new discrete diffusion process, transition kernel, or likelihood estimator. 
Instead, we study how a pretrained AR model can be converted into a DLM without relearning its internal language representations from scratch. 
This changes the optimization problem from full diffusion pretraining to representation-preserving mechanism adaptation: the DLM must learn to denoise under bidirectional attention and any-order generation, while remaining anchored to the semantic coordinate system already formed by AR pretraining.
The masked-denoising objective used in the main method is given in \Cref{sec:method_mdm}.

\subsection{Autoregressive-to-diffusion language model conversion}
\label{app:rw_ar_to_dlm}

The closest line of work studies how pretrained autoregressive language models can be adapted into diffusion language models. 
Gong et al. propose one of the first systematic AR$\rightarrow$DLM adaptation recipes, showing that GPT- and LLaMA-style checkpoints can be converted into DiffuGPT and DiffuLLaMA through continued diffusion training~\citep{gong2025scalingdiffusionlanguagemodels}. 
A key ingredient in this conversion is reconciling the next-token indexing convention of AR models with the position-wise denoising convention of masked diffusion, for example through shift operations during training and sampling. 
Subsequent large-scale DLMs further exploit AR initialization. 
Dream uses AR-based initialization together with diffusion-specific training strategies for large-scale diffusion language modeling~\citep{dream2025}, while Dream-Coder adapts pretrained AR checkpoints to masked diffusion for code generation~\citep{dreamcoder2025}. 
Efficient-DLM studies the conversion problem through attention and objective design, emphasizing that preserving favorable properties of the AR checkpoint is important for both quality and inference efficiency~\citep{efficientdlm2025}. 
Relatedly, Any-Order GPT connects masked diffusion with any-order autoregressive generation in a decoder-only framework, further highlighting that the distinction between AR and DLMs is partly a distinction between generation mechanisms rather than model families~\citep{xue2025anyordergpt}. 
Code-oriented diffusion models such as DiffuCoder provide additional evidence that masked diffusion can be competitive in execution-based code-generation settings~\citep{diffucoder2025}.

These works establish that AR checkpoints are valuable starting points for DLM training. 
However, existing conversion methods primarily reuse AR parameters and then adapt the model through objective-level, attention-level, masking-level, or sampling-level changes. 
They do not explicitly constrain the converted DLM to preserve the internal representation geometry of the AR model throughout training. 
Our method adds this missing constraint. 
We keep a frozen AR teacher, initialize the DLM student from the same checkpoint, switch the student from causal to bidirectional attention, and align the student's hidden states to the teacher's layer-wise representations during denoising training. 
Thus, our method does not merely initialize from an AR model; it uses the AR model as a persistent representational anchor.
The corresponding conversion procedure in our work is defined in \Cref{sec:method_models} and evaluated under matched conditions in \Cref{sec:results_alignment,sec:results_scaling}.

\begin{table}[t]
\centering
\small
\setlength{\tabcolsep}{4pt}
\caption{
Comparison with closely related AR-to-DLM conversion and representation-alignment methods.
Existing AR-to-DLM methods reuse pretrained AR parameters, but do not explicitly preserve the AR model's hidden-state geometry during denoising training.
REPA-style methods use representation alignment, but typically align diffusion models to external encoders rather than to the same AR model being converted.
}
\label{tab:app_related_comparison}
\resizebox{\textwidth}{!}{%
\begin{tabular}{lcccc}
\toprule
Method 
& Starts from AR LM 
& DLM target 
& Hidden alignment 
& Same-architecture AR teacher \\
\midrule
Gong et al.~\citep{gong2025scalingdiffusionlanguagemodels}
& Yes 
& Yes 
& No 
& No \\
Dream~\citep{dream2025}
& Yes 
& Yes 
& No 
& No \\
Dream-Coder~\citep{dreamcoder2025}
& Yes 
& Yes 
& No 
& No \\
Efficient-DLM~\citep{efficientdlm2025}
& Yes 
& Yes 
& No 
& No \\
Any-Order GPT~\citep{xue2025anyordergpt}
& Yes 
& Yes 
& No 
& No \\
REPA~\citep{yu2025repa}
& No 
& Yes 
& Yes 
& No \\
\textbf{Ours}
& Yes 
& Yes 
& Yes 
& Yes \\
\bottomrule
\end{tabular}
}
\end{table}

\subsection{Any-order generation, iterative decoding, and path planning}
\label{app:rw_any_order}

Our work is also connected to prior studies of generation order. 
Autoregressive language models are usually trained with a fixed left-to-right factorization, but this factorization is only one possible ordering of the joint distribution. 
XLNet introduced permutation language modeling, showing that autoregressive pretraining can incorporate bidirectional context by optimizing over multiple factorization orders~\citep{yang2019xlnet}. 
More generally, any-order autoregressive models study training and inference when the generation order is allowed to vary rather than being fixed in advance~\citep{shih2022traininginferenceanyorderautoregressive}. 
These works support the view that generation order is a modeling mechanism rather than an intrinsic property of the data distribution.

Iterative masked decoding provides another precursor to masked diffusion generation. 
Mask-Predict generates sequences by repeatedly predicting masked positions and remasking low-confidence tokens, offering an early non-autoregressive sequence-generation procedure based on masked language models~\citep{ghazvininejad2019maskpredict}. 
MaskGIT later demonstrated the effectiveness of confidence-based iterative masked generation in visual token models~\citep{Chang_2022_CVPR}. 
Modern masked diffusion language models can be viewed as probabilistic successors to these iterative refinement methods: they define an explicit corruption process, train a denoiser over partially masked sequences, and sample by progressively resolving uncertainty across positions.

Recent work further shows that the choice of denoising path and token order can substantially affect DLM sampling and training. 
P2 studies path planning for masked diffusion sampling and shows that non-uniform remasking and update schedules can improve generation quality without changing the denoising model~\citep{peng2025pathplanningmaskeddiffusion}. 
PAPL extends this view to training by incorporating planner-aware path learning into diffusion language model optimization~\citep{peng2026planner}. 
In this paper, we use PAPL as part of the default DLM training recipe for both the baseline and representation-aligned models. 
Therefore, the comparisons isolate the contribution of representation alignment rather than path-planning improvements. 
Conceptually, this line of work reinforces our central premise: if generation order and denoising paths are mechanisms that can be changed after pretraining, then the key question is whether the underlying language representations can be preserved while the mechanism changes.
The role of PAPL in our experimental protocol is specified in \Cref{sec:setup,app:path_loss}; because PAPL is shared across baselines and aligned models, it does not define the comparison axis.

\subsection{Representation alignment and feature preservation}
\label{app:rw_repr_alignment}

Representation alignment has recently emerged as an effective way to accelerate generative model training. 
REPA aligns the hidden states of diffusion or flow-based transformers to representations from strong pretrained visual encoders, showing that generative training can be bottlenecked by slow representation learning rather than by denoising alone~\citep{yu2025repa}. 
Follow-up work studies which components of the target representation are most useful, whether global information or spatial structure matters most, and whether diffusion transformers can provide representation guidance without external encoders~\citep{singh2025irepa,wu2025representation,jiang2025sra}. 
Together, these methods suggest that explicitly constraining intermediate representations can improve the efficiency and stability of generative training.

Our setting differs in both the teacher and the purpose of alignment. 
Prior representation-alignment methods typically align a generative model to an external encoder, often in the vision domain. 
The teacher and student may have different architectures, modalities, or training objectives, so alignment acts as a form of semantic feature distillation. 
In contrast, our teacher is the exact AR model being converted: it has the same tokenizer, transformer architecture, hidden dimensionality, and pretrained weights as the DLM student before conversion. 
The student differs only in the generation mechanism induced by bidirectional attention and masked denoising. 
Thus, alignment is not used to import features from an external model; it is used to preserve the internal coordinate system of the pretrained AR model while learning a new denoising mechanism.

This distinction is central to our interpretation. 
If AR pretraining has already learned useful semantic and syntactic representations, then AR$\rightarrow$DLM conversion should not require relearning those representations from scratch. 
The remaining problem is to adapt the model so that these representations can support any-order denoising rather than left-to-right next-token prediction. 
Layer-wise representation alignment directly implements this view: it anchors the DLM student to the frozen AR teacher while the diffusion loss trains the student to operate under masked bidirectional generation. 
The resulting method combines the practical benefits of AR initialization with an explicit constraint that preserves the AR model's hidden-state geometry throughout conversion.
The same-architecture AR-teacher version of this idea is formalized in \Cref{sec:method_alignment} and ablated in \Cref{sec:ablations}.

\section{Experimental Details}
\label{app:exp_details}

This appendix gives the implementation details for the experimental protocol summarized in \Cref{sec:setup} and used in \Cref{sec:results,sec:ablations}. Unless otherwise specified, all experiments use the same model initialization, data preprocessing, masked denoising objective, path-planning loss, optimization configuration, and decoding protocol. Ablations vary only the factor explicitly stated.

\subsection{Model Conversion}
\label{app:model_conversion}

We convert a pretrained autoregressive language model into a masked diffusion language model by changing the attention pattern from causal to bidirectional during masked diffusion training. The transformer architecture is otherwise unchanged: the model remains a decoder-only Qwen-style transformer with rotary position embeddings, RMSNorm, MLP blocks, and a tied input embedding/output language-modeling head. The original AR model is used as a frozen teacher, and the DLM student is initialized from the same checkpoint.
This implements the two-model construction in \Cref{sec:method_models}.

In the implementation, the bidirectional attention mode is activated only for masked diffusion training. When the forward pass receives a mask ratio, the model sets \texttt{is\_causal=False}; otherwise, it preserves the standard causal mode. Diffusion generation uses full-sequence non-causal forward passes and does not use KV caching.

\begin{lstlisting}[style=pyalg, caption={Causal-to-bidirectional conversion.}, label={lst:causal_to_bidir}]
# In Qwen*ForCausalLM.forward
if mask_ratio is not None:
    is_causal = False
else:
    is_causal = True

outputs = self.model(
    input_ids=input_ids,
    attention_mask=attention_mask,
    is_causal=is_causal,
    output_hidden_states=output_hidden_states,
)
\end{lstlisting}

Following AR-to-DLM adaptation, we use a one-token shift convention to reconcile next-token prediction with position-wise masked denoising. During training, labels are shifted left and hidden states are sliced accordingly. During sampling, logits are shifted so that diffusion predictions remain consistent with the inherited AR indexing convention. The same shift convention is used for both the baseline and representation-aligned models.

\subsection{Data and Preprocessing}
\label{app:data_preprocessing}

Training uses the Nemotron-SFT-Code corpus stored as local Parquet files. The training stream is loaded in streaming mode. Each example is read from the \texttt{text} field, appended with an EOS token, and tokenized with \texttt{add\_special\_tokens=False}. Examples longer than the maximum sequence length are split into chunks of length 4096, while shorter examples are kept and packed. When packing is enabled, multiple examples are concatenated into packed rows with reset position ids, so that sequence boundaries are preserved. No additional filtering is applied.
This preprocessing supports the data regimes described in \Cref{sec:setup}, including the tiny-data comparison in \Cref{sec:results_tiny}.

\begin{table}[t]
\centering
\small
\caption{Data preprocessing configuration.}
\label{tab:app_data_config}
\begin{tabular}{ll}
\toprule
Setting & Value \\
\midrule
Dataset format & Local Parquet files \\
Number of files & 78 \\
Number of rows & $76{,}447{,}340$ \\
Text field & \texttt{text} \\
Tokenization & append EOS, \texttt{add\_special\_tokens=False} \\
Maximum sequence length & 4096 \\
Long examples & split into 4096-token chunks \\
Short examples & kept and packed \\
Packing & concatenate examples with reset position ids \\
Shuffle seed & 42 \\
Shuffle buffer & 10000 \\
Filtering & none \\
\bottomrule
\end{tabular}
\end{table}

The main assets used in the experiments are licensed or governed as follows: the Qwen3 checkpoints are released under Apache 2.0; the Nemotron-SFT-Code training corpus is governed by NVIDIA's data-access terms for model training; HumanEval is released under the original MIT license from OpenAI; MBPP is released under CC BY 4.0 via the Google Research dataset release; and the EvalPlus HumanEval+/MBPP+ releases are Apache 2.0.

\subsection{Masked Diffusion Objective}
\label{app:masked_objective}

This section implements the main denoising loss in \Cref{eq:diff_loss}.
The DLM student is trained with a masked denoising objective. For each original sequence, we sample a mask ratio
\begin{equation}
    r \sim \mathrm{Uniform}\left(\frac{1}{500}, 1-\frac{1}{500}\right).
\end{equation}
Conditioned on this ratio, each token position is independently masked with probability $r$. Thus, the implementation uses Bernoulli masking rather than selecting exactly $\lfloor rn \rceil$ masked positions. The cross-entropy loss is computed only on masked positions, while unmasked positions and packed-sequence boundary positions are assigned \texttt{IGNORE\_INDEX}.

\begin{lstlisting}[style=pyalg, caption={Masked diffusion corruption.}, label={lst:masking}]
mask_ratio = torch.rand(1).clamp(1 / 500, 1 - 1 / 500)
mask = torch.rand_like(input_ids.float()) < mask_ratio

labels = input_ids.clone()
input_ids[mask] = mask_token_id
labels[input_ids != mask_token_id] = IGNORE_INDEX
\end{lstlisting}

Let $M$ denote the set of masked and shift-valid positions. The masked denoising loss is
\begin{equation}
\mathcal{L}_{\mathrm{mdm}}
=
\frac{1}{|M|}
\sum_{i\in M}
\mathrm{CE}\!\left(p_{\theta}(\cdot\mid \tilde{x})_i, x_i\right),
\end{equation}
where $\tilde{x}$ is the corrupted input sequence. In implementation, token losses are summed over valid masked positions and normalized by the number of valid masked tokens.

\subsection{Representation Alignment}
\label{app:repr_alignment_details}

For representation alignment, we construct a frozen teacher by deep-copying the initialized AR model. The teacher is placed in evaluation mode and all teacher parameters are frozen. During training, the teacher consumes the clean sequence under causal attention, while the student consumes the masked sequence under bidirectional attention. Teacher features are computed under \texttt{torch.no\_grad()}.
This section implements the alignment loss in \Cref{eq:align_loss}, including which hidden states and token positions are included.

We align the hidden-state tuple returned by the model, including the embedding hidden state, block outputs, and final post-normalization hidden state. By default, all returned hidden states are aligned. Layer ablations select contiguous thirds of this hidden-state tuple. Alignment is applied only on masked and shift-valid positions, matching the positions that contribute to the denoising loss.

For cosine alignment, hidden states are normalized along the feature dimension and the loss is computed as one minus the mean cosine similarity:
\begin{equation}
\mathcal{L}_{\mathrm{align}}
=
1 -
\frac{1}{|\mathcal{H}|}
\sum_{h\in \mathcal{H}}
\frac{1}{|M|}
\sum_{i\in M}
\left\langle
\frac{h_{\mathrm{D},i}}{\|h_{\mathrm{D},i}\|_2},
\frac{\mathrm{sg}(h_{\mathrm{AR},i})}{\|\mathrm{sg}(h_{\mathrm{AR},i})\|_2}
\right\rangle,
\end{equation}
where $\mathcal{H}$ is the selected set of hidden states, $M$ is the masked and shift-valid position set, and $\mathrm{sg}(\cdot)$ denotes stop-gradient.

\begin{lstlisting}[style=pyalg, caption={Layer-wise representation alignment.}, label={lst:repr_align}]
with torch.no_grad():
    teacher_outputs = teacher(
        input_ids=clean_input_ids,
        is_causal=True,
        output_hidden_states=True,
    )

student_outputs = student(
    input_ids=masked_input_ids,
    is_causal=False,
    output_hidden_states=True,
)

loss_mask = labels != IGNORE_INDEX
repr_loss = cosine_distance(
    student_outputs.hidden_states,
    teacher_outputs.hidden_states,
    loss_mask=loss_mask,
)

loss = mdm_loss + path_loss + lambda_repr * repr_loss
\end{lstlisting}

The default alignment weight is $\lambda_{\mathrm{repr}}=10$. Unless otherwise specified, all aligned models use cosine alignment over all selected hidden states.

\subsection{Path-Planning Loss}
\label{app:path_loss}

All DLM runs, including both the baseline and representation-aligned models, include the same path-planning auxiliary loss. We therefore treat this loss as part of the default DLM training recipe rather than as a separate method-specific component. This ensures that comparisons isolate the effect of \nameshort.
This term is included in both baseline and aligned models in \Cref{sec:setup}, so the main comparisons isolate $\mathcal{L}_{\mathrm{align}}$.

For Qwen3-based runs, the path-planning loss reweights the masked-token cross-entropy by the detached model confidence:
\begin{equation}
\mathcal{L}_{\mathrm{path}}
=
\frac{1}{|M|}
\sum_{i\in M}
\exp(-\ell_i)_{\mathrm{sg}} \, \ell_i,
\end{equation}
where $\ell_i$ is the token-level cross-entropy loss at masked position $i$, and the confidence weight is stop-gradiented. The total training loss is
\begin{equation}
\mathcal{L}
=
\mathcal{L}_{\mathrm{mdm}}
+
\mathcal{L}_{\mathrm{path}}
+
\lambda_{\mathrm{repr}}\mathcal{L}_{\mathrm{align}}.
\end{equation}
For baseline DLM conversion, the representation-alignment term is omitted:
\begin{equation}
\mathcal{L}_{\mathrm{baseline}}
=
\mathcal{L}_{\mathrm{mdm}}
+
\mathcal{L}_{\mathrm{path}}.
\end{equation}

\subsection{Optimization}
\label{app:optimization}

All models are optimized with AdamW. Unless otherwise specified, we use the same optimization configuration across baseline and representation-aligned runs.
This optimization protocol is the fixed-budget setting used in \Cref{sec:results_alignment,sec:results_scaling}.

\begin{table}[t]
\centering
\small
\caption{Default optimization configuration.}
\label{tab:app_optimization}
\begin{tabular}{ll}
\toprule
Setting & Value \\
\midrule
Optimizer & AdamW \\
Learning rate & $3\times 10^{-4}$ \\
Adam betas & $(0.9, 0.95)$ \\
Adam epsilon & $10^{-8}$ \\
Weight decay & $0.01$ \\
Parameter groups & single group for all trainable parameters \\
LR schedule & cosine decay \\
Warmup ratio & $0.001$ \\
Minimum learning rate & $3\times 10^{-6}$ \\
Gradient clipping & $1.0$ \\
Precision & bfloat16 \\
Maximum sequence length & 4096 \\
Per-device batch size & 6 \\
Gradient accumulation & 1 \\
Global batch size & 96 sequences \\
Number of GPUs & 16 \\
Parallelism & DDP \\
Attention backend & FlashAttention-2 \\
Gradient checkpointing & disabled \\
FSDP / DeepSpeed & disabled \\
\bottomrule
\end{tabular}
\end{table}

\subsection{Decoding and Evaluation}
\label{app:evaluation_details}

We evaluate code generation on HumanEval~\citep{chen2021evaluating}, MBPP~\citep{austin2021program}, and the HumanEval+/MBPP+ variants from EvalPlus~\citep{liu2023evalplus}. HumanEval uses the canonical prompt from the benchmark. MBPP uses the task description together with the first three tests and ends with a Python code block. All evaluations are zero-shot.
This evaluation protocol underlies the main comparisons in \Cref{sec:results_public} and the ablations in \Cref{sec:ablations}.

For each problem, we generate 10 samples and compute execution-based pass@1 and pass@10 using the standard code-evaluation pipeline. The same decoding configuration is used for all methods being compared.

\begin{table}[t]
\centering
\small
\caption{Default decoding and evaluation configuration.}
\label{tab:app_eval_config}
\begin{tabular}{ll}
\toprule
Setting & Value \\
\midrule
Benchmarks & HumanEval, HumanEval+, MBPP, MBPP+ \\
Prompting & zero-shot \\
Samples per problem & 10 \\
Metric & execution-based pass@1/pass@10 \\
Sampler & P2-self, \texttt{alg=p2} \\
Sampling steps & 128 \\
Maximum new tokens & 128 \\
Temperature & 0.8 \\
Top-$k$ & 200 \\
Algorithm temperature & 0.5 \\

\bottomrule
\end{tabular}
\end{table}

The P2 sampler keeps prompt tokens fixed and iteratively remasks low-confidence variable positions. At each step, the model predicts the full sequence, samples variable positions, and then remasks a fraction of the lowest-confidence generated tokens according to the sampling schedule.

\begin{lstlisting}[style=pyalg, caption={Evaluation command template.}, label={lst:eval_command}]
accelerate launch --num_processes 4 eval.py \
  --model custom_coder \
  --model_args "pretrained=<checkpoint>,max_new_tokens=128,steps=128,temperature=0.8,alg=p2" \
  --tasks humaneval --num_fewshot 0 --batch_size 10 \
  --output_path evals_results/humaneval-ns0 --log_samples \
  --confirm_run_unsafe_code
\end{lstlisting}

\subsection{Ablation Protocols}
\label{app:ablation_protocols}

All ablations use the same AR$\rightarrow$DLM conversion pipeline unless otherwise stated. In particular, the model is initialized from the same AR checkpoint, trained with the same masked denoising objective and path-planning loss, and evaluated with the same decoding configuration. Each ablation varies only the listed factor.
These protocols correspond to the ablation results in \Cref{sec:ablations}.

\begin{table}[t]
\centering
\small
\caption{Ablation protocols.}
\label{tab:app_ablation_protocol}
\begin{tabular}{lll}
\toprule
Ablation & Variants & Changed factor \\
\midrule
Alignment metric & L2, cosine & representation loss \\
Alignment weight & $1, 5, 10, 20$ & $\lambda_{\mathrm{repr}}$ \\
Aligned layers & lower, middle, upper, all & selected hidden states \\
Freezing & none, embeddings/LM head/MLP & trainable parameter blocks \\
Data size & full, reduced & training stream size \\
\bottomrule
\end{tabular}
\end{table}

For the alignment-metric ablation, L2 alignment uses mean-squared error between student and teacher hidden states, while cosine alignment normalizes hidden states along the feature dimension and matches their directions. For the layer ablation, we partition the hidden-state tuple into three contiguous groups and apply the representation loss only to one group at a time. The default setting aligns all hidden states with cosine distance and $\lambda_{\mathrm{repr}}=10$.

\begin{table}[]
\centering
\small
\caption{Freezing configuration.}
\label{tab:app_freezing}
\begin{tabular}{lll}
\toprule
Variant & Frozen parameters & Trainable parameters \\
\midrule
No freezing & none & all parameters \\
Embedding/MLP freezing & \texttt{embed\_tokens}, \texttt{lm\_head}, \texttt{mlp} & attention and normalization layers \\
\bottomrule
\end{tabular}
\end{table}

\subsection{Freezing Protocol}
\label{app:freezing_protocol}

To test whether AR$\rightarrow$DLM conversion requires updating the full network, we freeze selected parameter blocks during representation-aligned training. Freezing is implemented by case-insensitive substring matching on parameter names. In the main freezing variant, parameters whose names contain \texttt{embed\_tokens}, \texttt{lm\_head}, or \texttt{mlp} are frozen. Attention layers and normalization layers remain trainable.
This protocol supports the efficiency result in \Cref{sec:results_freezing}.

This freezing protocol is designed to preserve most of the pretrained representational and feed-forward computation while allowing the attention mechanism and normalization statistics to adapt to bidirectional denoising.

\section{Limitations}
\label{sec:limitations}
Our study is limited to same-architecture AR$\rightarrow$DLM conversion on Qwen3 decoder-only checkpoints and code-generation benchmarks. The gains we report may not transfer unchanged to other model families, modalities, or downstream tasks, and the method still depends on access to a strong pretrained AR teacher and substantial training compute.

\section{Broader Impacts}
\label{sec:broader_impacts}
This work can lower the compute barrier for training competitive diffusion language models, which may make efficient generation research more accessible. Cheaper conversion of pretrained language models can also make code synthesis and other generative capabilities easier to scale for harmful or deceptive uses, so deployment should follow ordinary model-stewardship practices. Because our contribution is a training method rather than a released model or dataset, we do not propose separate release restrictions.

\let\section\appendixsection
\let\subsection\appendixsubsection

\newpage

\section*{NeurIPS Paper Checklist}
\begin{enumerate}

\item {\bf Claims}
    \item[] Question: Do the main claims made in the abstract and introduction accurately reflect the paper's contributions and scope?
    \item[] Answer: \answerYes{}
    \item[] Justification: The abstract and introduction accurately describe the paper's scope: same-architecture AR$\rightarrow$DLM conversion with layer-wise representation alignment, evaluated on Qwen3 checkpoints and code-generation benchmarks. The claims are consistent with the experiments and ablations reported in \Cref{sec:results,sec:ablations}.

\item {\bf Limitations}
    \item[] Question: Does the paper discuss the limitations of the work performed by the authors?
    \item[] Answer: \answerYes{}
    \item[] Justification: The revised paper now includes a dedicated \Cref{sec:limitations} section. It states the main scope limits clearly: same-architecture Qwen3 conversion, code-generation benchmarks, dependence on a strong pretrained AR teacher, and substantial training compute.

\item {\bf Theory assumptions and proofs}
    \item[] Question: For each theoretical result, does the paper provide the full set of assumptions and a complete (and correct) proof?
    \item[] Answer: \answerNA{}
    \item[] Justification: The paper does not contain new theorems, lemmas, or formal proofs; it presents an empirical method with explicit equations and implementation details instead.

    \item {\bf Experimental result reproducibility}
    \item[] Question: Does the paper fully disclose all the information needed to reproduce the main experimental results of the paper to the extent that it affects the main claims and/or conclusions of the paper (regardless of whether the code and data are provided or not)?
    \item[] Answer: \answerYes{}
    \item[] Justification: The paper specifies the model family, checkpoint source, training data, preprocessing, optimization hyperparameters, decoding setup, and ablation protocol in \Cref{sec:setup,app:exp_details}. Those details are sufficient for another group to reproduce the main experiments in principle.

\item {\bf Open access to data and code}
    \item[] Question: Does the paper provide open access to the data and code, with sufficient instructions to faithfully reproduce the main experimental results, as described in supplemental material?
    \item[] Answer: \answerNo{}
    \item[] Justification: The manuscript gives detailed reproduction instructions, but it does not yet provide a public release of the training data or converted model checkpoints from this paper. Reproducibility is therefore documented, but open access to all code/data assets is not provided in the paper itself.

\item {\bf Experimental setting/details}
    \item[] Question: Does the paper specify all the training and test details (e.g., data splits, hyperparameters, how they were chosen, type of optimizer) necessary to understand the results?
    \item[] Answer: \answerYes{}
    \item[] Justification: The core paper and appendix specify the model families, training corpus, preprocessing, optimization hyperparameters, decoding configuration, and evaluation protocol. The ablation appendix also records the factors changed in each study.

\item {\bf Experiment statistical significance}
    \item[] Question: Does the paper report error bars suitably and correctly defined or other appropriate information about the statistical significance of the experiments?
    \item[] Answer: \answerNo{}
    \item[] Justification: The reported benchmark tables and plots give point estimates only; the manuscript does not include error bars, confidence intervals, or multi-seed significance tests for the main results.

\item {\bf Experiments compute resources}
    \item[] Question: For each experiment, does the paper provide sufficient information on the computer resources (type of compute workers, memory, time of execution) needed to reproduce the experiments?
    \item[] Answer: \answerNo{}
    \item[] Justification: The appendix reports that training used 16 GPUs with DDP and gives the batch/optimizer setup, but it does not yet specify the accelerator model, memory footprint, or wall-clock time for the main runs.
    
\item {\bf Code of ethics}
    \item[] Question: Does the research conducted in the paper conform, in every respect, with the NeurIPS Code of Ethics \url{https://neurips.cc/public/EthicsGuidelines}?
    \item[] Answer: \answerYes{}
    \item[] Justification: The research uses public or properly licensed model and benchmark assets, involves no human subjects or sensitive data collection, and does not identify any deviation from the NeurIPS Code of Ethics. The paper's conduct is consistent with the ethics guidelines as written.

\item {\bf Broader impacts}
    \item[] Question: Does the paper discuss both potential positive societal impacts and negative societal impacts of the work performed?
    \item[] Answer: \answerYes{}
    \item[] Justification: The revised paper now includes a short broader-impacts section that discusses both benefits from lower compute cost and risks from cheaper scaling of generative code capabilities. It also states that the paper does not propose gated release or usage restrictions because it is a training method rather than a released model.
    
\item {\bf Safeguards}
    \item[] Question: Does the paper describe safeguards that have been put in place for responsible release of data or models that have a high risk for misuse (e.g., pre-trained language models, image generators, or scraped datasets)?
    \item[] Answer: \answerNA{}
    \item[] Justification: The paper does not release a new model or dataset with elevated misuse risk, so there is no separate release mechanism to safeguard here. If a future checkpoint release is added, it should be accompanied by a dedicated safeguards statement.

\item {\bf Licenses for existing assets}
    \item[] Question: Are the creators or original owners of assets (e.g., code, data, models), used in the paper, properly credited and are the license and terms of use explicitly mentioned and properly respected?
    \item[] Answer: \answerYes{}
    \item[] Justification: The paper now states the relevant licenses/terms for the main assets: Qwen3 checkpoints (Apache 2.0), Nemotron-SFT-Code (NVIDIA model-training data agreement), HumanEval (MIT), MBPP (CC BY 4.0), and EvalPlus HumanEval+/MBPP+ (Apache 2.0). Each asset is also cited in the bibliography.

\item {\bf New assets}
    \item[] Question: Are new assets introduced in the paper well documented and is the documentation provided alongside the assets?
    \item[] Answer: \answerNA{}
    \item[] Justification: The manuscript does not introduce a newly released dataset, benchmark, or model package in its current form.

\item {\bf Crowdsourcing and research with human subjects}
    \item[] Question: For crowdsourcing experiments and research with human subjects, does the paper include the full text of instructions given to participants and screenshots, if applicable, as well as details about compensation (if any)? 
    \item[] Answer: \answerNA{}
    \item[] Justification: The paper does not involve crowdsourcing or human-subject experiments.

\item {\bf Institutional review board (IRB) approvals or equivalent for research with human subjects}
    \item[] Question: Does the paper describe potential risks incurred by study participants, whether such risks were disclosed to the subjects, and whether Institutional Review Board (IRB) approvals (or an equivalent approval/review based on the requirements of your country or institution) were obtained?
    \item[] Answer: \answerNA{}
    \item[] Justification: The paper does not involve human subjects, so IRB approval or equivalent review is not applicable.

\item {\bf Declaration of LLM usage}
    \item[] Question: Does the paper describe the usage of LLMs if it is an important, original, or non-standard component of the core methods in this research? Note that if the LLM is used only for writing, editing, or formatting purposes and does \emph{not} impact the core methodology, scientific rigor, or originality of the research, declaration is not required.
    \item[] Answer: \answerYes{}
    \item[] Justification: The core method and evaluation are explicitly built around LLM checkpoints and benchmarks, especially Qwen3-based AR/DLM conversion. The paper documents that usage throughout the method and experimental sections.

\end{enumerate}

\end{document}